\documentclass[11pt]{article}

\usepackage[final]{acl}

\usepackage{times}
\usepackage{latexsym}
\usepackage{booktabs}
\usepackage{multirow}
\usepackage{graphicx}
\usepackage{enumitem}
\usepackage{amsmath}
\usepackage{listings}
\usepackage{mdframed}
\usepackage{xcolor}
\usepackage{verbatim}
\usepackage[T1]{fontenc}

\usepackage[utf8]{inputenc}

\usepackage{microtype}

\usepackage{inconsolata}

\hypersetup{
  pdftitle={How Well Can Modern LLMs Act as Agent Cores in Radiology Environments?},
  pdfauthor={Qiaoyu Zheng, Chaoyi Wu, Weike Zhao, Pengcheng Qiu, Lisong Dai, Ya Zhang, Yanfeng Wang, Weidi Xie}
}

\title{How Well Can Modern LLMs Act as Agent Cores in Radiology Environments?}

\author{
 \textbf{Qiaoyu Zheng\textsuperscript{1,*}},
 \textbf{Chaoyi Wu\textsuperscript{1,*}},
 \textbf{Weike Zhao\textsuperscript{1}},
 \\[3pt]
 \textbf{Pengcheng Qiu\textsuperscript{1}},
 \textbf{Lisong Dai\textsuperscript{2}},
 \textbf{Ya Zhang\textsuperscript{1}},
 \\[3pt]
 \textbf{Yanfeng Wang\textsuperscript{1,$\dag$}},
 \textbf{Weidi Xie\textsuperscript{1,$\dag$}},
 \\[3pt]
 \textsuperscript{1}Shanghai Jiao Tong University, Shanghai, China
 \\[3pt]
 \textsuperscript{2}Department of Radiology, Renmin Hospital of Wuhan University, Wuhan, China
 \\[3pt]
 \texttt{\{three-world,wangyanfeng622,weidi\}@sjtu.edu.cn}
 \\[3pt]
 \href{https://github.com/MAGIC-AI4Med/RadABench}{https://github.com/MAGIC-AI4Med/RadABench}
 \\[3pt]
 \textsuperscript{*}Equal contribution. \quad \textsuperscript{$\dag$}Corresponding authors
}

\begin{document}
\maketitle
\begin{abstract}
Radiology, with its heterogeneous modalities, anatomies, and evolving protocols, is a natural yet high-stakes testbed for agentic AI. As LLMs grow more capable and tool ecosystems such as MCP and Agent Skills more complex, their capability boundaries in radiology remain unclear. We introduce \textbf{RadA-BenchPlat}, a two-layer benchmark spanning idealized reasoning and real-world execution: a \textit{synthetic layer} with 2.2k clinician-verified records, 24.2k QA pairs, and 10 tool categories under diverse availability settings; and a \textit{real-environment layer} pairing 165 2D/3D cases with executable tools for grounding, diagnosis, and report generation. We find that high completion rates under idealized settings do not carry over to real environments, where the best agents trail their upper bound by $\sim$15\% and drop to 0.327 on long-chain tasks. Prompting strategies (few-shot exemplars, multi-agent coordination) and AutoTB (on-the-fly synthesis of missing tools) mitigate execution drift and breakdowns in difficult scenarios—lifting execution success to 94.5\%—yet remain not fully reliable. Overall, our findings suggest that radiology agents are no longer limited to conceptual demonstrations, but are beginning to show potential for real-world applications.
\end{abstract}
\section{Introduction}

\begin{figure}[t]
    \centering
    \includegraphics[width=\linewidth]{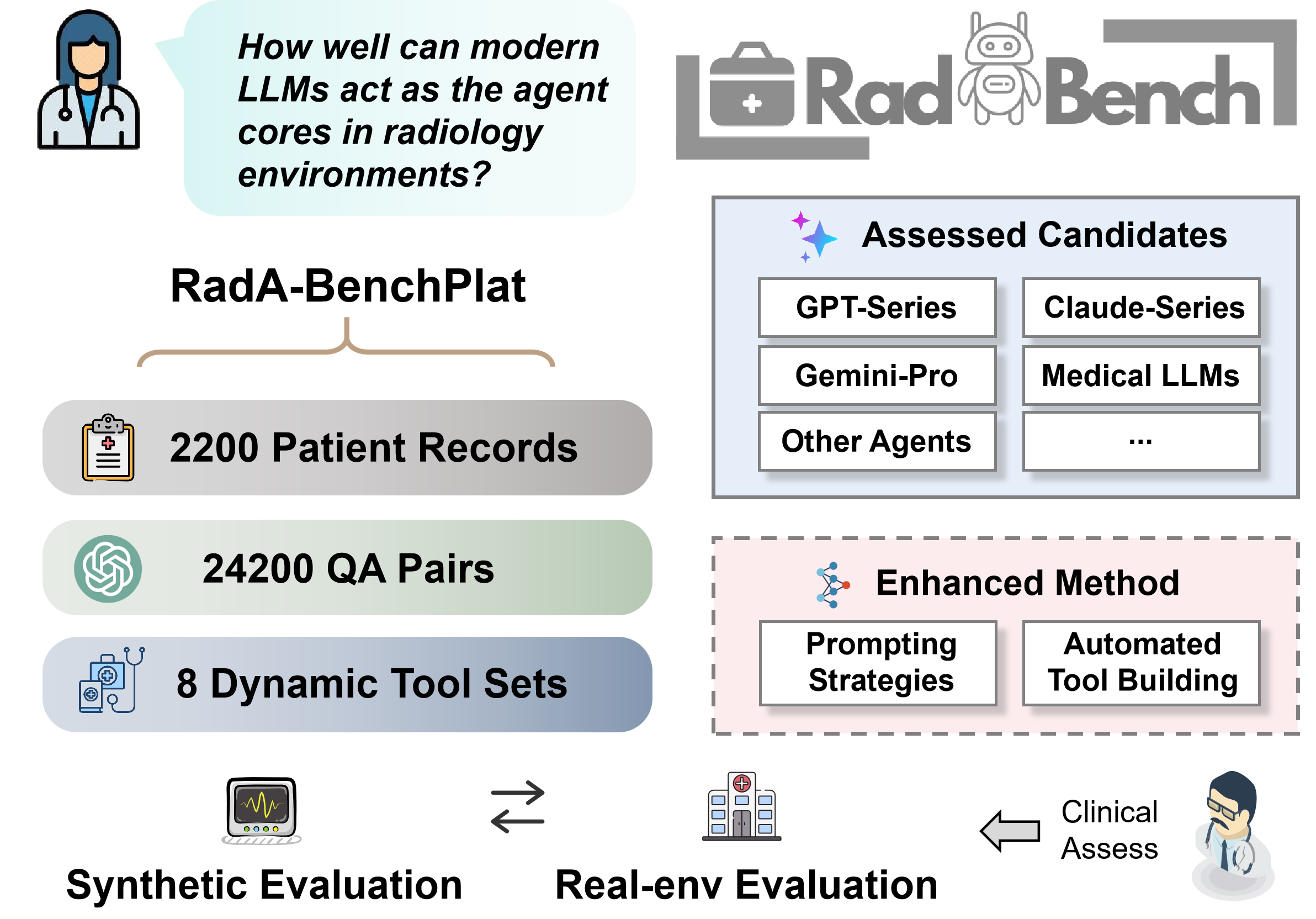}
    \caption{\textbf{Overview.} Motivation and components of the synthetic and real-environment benchmark layers.}
    \label{fig:Overview}
\end{figure}

Recent efforts have introduced ``radiology generalist'' models designed to handle diverse imaging tasks within a unified framework~\cite{wu2023towards, li2023llavamed, liu2023radiology, zhang2023pmc}. However, these monolithic systems still struggle with radiological complexity across heterogeneous imaging modalities, anatomical regions, pathologies, and evolving diagnostic protocols. In parallel, LLM-powered agent systems~\cite{tang2024medagents, jin2024agentmd, qin2023toolllm} reframe complex problem solving as tool-mediated planning and execution, where an LLM interprets requests, selects specialized tools, and synthesizes intermediate outputs. This paradigm is increasingly relevant as agent ecosystems move beyond static tool lists toward heterogeneous infrastructures, such as Model Context Protocol (MCP) servers~\cite{Zhu2026MedMCPCalcBL} and Agent Skills~\cite{Xu2026AnES}, where capabilities, interfaces, and dependencies vary substantially across environments. As modern LLMs become more agentic, radiology offers a high-stakes setting for examining whether they can reliably coordinate complex tool chains; however, their capability boundaries, robustness, and danger zones on radiology tasks remain insufficiently and systematically studied.

We therefore focus on two questions. \textbf{First}, if individual tool quality is abstracted away, how well can current LLMs understand radiological contexts, translate clinical queries into executable workflows, and invoke tools across multi-step tasks? \textbf{Second}, in realistic radiology settings with executable tools, what level of task completion and clinical utility can current agent systems achieve?

To address these questions, we introduce \textbf{RadA-BenchPlat}, a benchmark with two complementary layers (Figure~\ref{fig:Overview}). \textbf{First}, the synthetic layer integrates patient records, task-related QA pairs, and simulated radiology tool sets: 2,200 records across 6 anatomical regions and 5 imaging modalities, cross-validated by three clinicians on 550 sampled cases; 11 task types yielding 24,200 QA pairs; and 10 tool categories organized into baseline, redundant, insufficient, and differentiated settings. \textbf{Second}, the real-environment layer uses 165 cases from 2D and 3D datasets for three representative tasks---organ/anomaly grounding, diagnosis with grounding clues, and report generation---and introduces executable radiology-model tools from $\text{M}^3$Builder~\cite{feng2025m}, nnU-Net~\cite{Isensee2020nnUNetAS}, and RP3D-DM~\cite{zheng2024large}.

Our synthetic evaluation reveals a central insight: current LLM agents can often outline plausible radiology workflows, but their idealized upper bound is limited by execution rather than reasoning alone. Reliability degrades when tasks require long-chain dependency tracking, precise I/O transfer, and grounded final synthesis; in high-difficulty settings, the best completion remains only 0.327. Nevertheless, few-shot workflow exemplars and multi-agent coordination alleviate long-chain control failures, while AutoTB, inspired by self-evolving agents that create tools on demand~\cite{cai2023large, wolflein2025llm}, addresses missing-capability cases by requesting missing tools in the synthetic layer and constructing them in the real-environment layer. Together, these results show that workflow-level scaffolding can raise the attainable ceiling.

Real-environment evaluation sharpens this finding: even the best-performing agentic systems remain about 15\% below the idealized setting, revealing a persistent gap between reasoning over tools and executing with them. AutoTB narrows this gap in regular and intermediate missing-tool scenarios (94.5\% and 86.8\% execution success), yet 48\% usage errors show that newly built tools still require robust integration. Expert review suggests practical value in routine and intermediate cases (4.2 and 3.6 above a 3-point threshold), while complex cases remain clinically fragile. These findings suggest that radiology agents are ready for controlled clinical-scenario testing, but their real-world effectiveness must still be carefully validated.

\section{Methods}
\begin{figure*}[!h]
    \centering
    \includegraphics[width=1\linewidth]{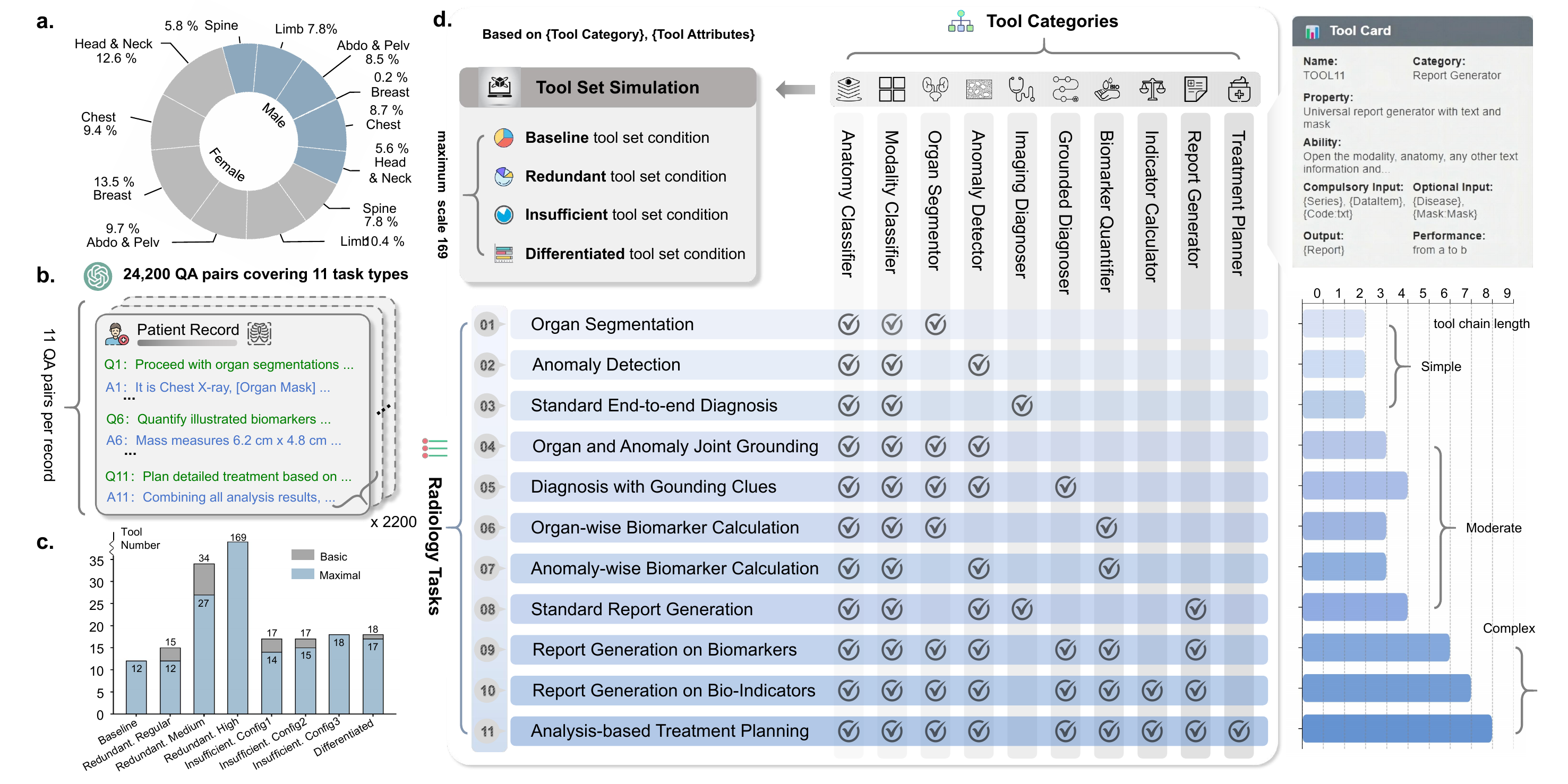}
    \caption{\textbf{Statistics in RadA-BenchPlat.}
    \textbf{a.} Distribution of patient sex across different anatomical regions. 
    \textbf{b.} Ten categories of tools and eleven types of radiology tasks, with checkmarks indicating tools required for task completion. Tool sets are simulated through various tool attribute combinations. \textbf{c.} 24,200 generated QA-pairs based on radiology task types using GPT-4. \textbf{d.} Number of tools across eight simulated tool set conditions.}
    \label{fig:EvalPlat}
\end{figure*}

RadA-BenchPlat is organized as two complementary layers. The synthetic layer builds an idealized environment with perfect simulated tools, which decouples failures of the agent core from failures of tool execution and stress-tests a broad range of scenarios that limited real data cannot cover. The real-environment layer instead measures what current agentic systems actually achieve on executable radiology tasks and what limits them in practice.

\subsection{Building RadA-BenchPlat}

\noindent \textbf{Patient Records.}
A ``patient record'' denotes a radiology-centric record containing demographics, medical history, annotated imaging data~(anatomy and pathology), and clinical findings. 
Records are categorized by \textbf{anatomy, modality, and disease}, covering 22 anatomy--modality combinations across 6 regions and 5 imaging modalities:
\vspace{-1.2mm}
\begin{quote}
    \textbf{Head \& Neck:} \{\textit{X-ray, CT, MRI, US}\}, \textbf{Chest:} \{\textit{X-ray, CT, MRI, US}\}, \\
     \textbf{Limb:} \{\textit{X-ray, CT, MRI, US}\}, \\ \textbf{Abd. \& Pelv.:} \{\textit{X-ray,  CT, MRI, US}\}, \\ 
    \textbf{Spine:} \{\textit{X-ray, CT, MRI}\}, \\
    \textbf{Breast:} \{\textit{Mammo., MRI, US}\}.
\end{quote}
\vspace{-1.2mm}
For each of the 22 anatomy--modality combinations, GPT-4 was used to generate 100 disease-centered records, yielding 2,200 patient records in total (22 $\times$ 100). Generation is driven by structured templates that constrain the output fields; the construction pipeline for records, QA pairs, and simulated tool sets is detailed in Appendix~\ref{sec:construction_pipeline}. These records were then carefully polished for more than one month by a radiologist with over 10 years of experience, with 97.32\% meeting information-validity standards and 96.73\% judged clinically consistent. To reduce single-annotator bias, three additional clinicians independently reviewed 550 randomly sampled cases across eight quality criteria. Their average positive percent agreement (PPA) ranges from 0.880 to 0.935 (Table~\ref{tab:clinician_validation_dataset}). Detailed rater backgrounds, criteria, and validation procedures are provided in Appendix~\ref{sec:clinician_validation}. Figure~\ref{fig:EvalPlat}a summarizes the demographic distributions: the dataset is generally sex-balanced except for female breast-related cases, and age, height, and weight follow clinically plausible ranges.

\begin{table*}[!t]
\centering
\caption{\textbf{Clinician validation of the synthetic dataset.} Each rater row reports the number of passing items among 550 randomly sampled cases. Avg PPA denotes the average positive percent agreement across the three raters.}
\label{tab:clinician_validation_dataset}
\scriptsize
\setlength{\tabcolsep}{3pt}
\resizebox{\textwidth}{!}{%
\begin{tabular}{lcccccccc}
\toprule
\textbf{Rater} & \textbf{Clinical Consist.} & \textbf{Info. Validity} & \textbf{Tool Necessity} & \textbf{No Leakage} & \textbf{Reasoning Depth} & \textbf{Ambiguity} & \textbf{Professionalism} & \textbf{Bias/Diversity} \\
\midrule
1 & 519 & 498 & 514 & 509 & 516 & 492 & 516 & 511 \\
2 & 514 & 489 & 509 & 501 & 512 & 483 & 506 & 506 \\
3 & 509 & 483 & 504 & 493 & 506 & 477 & 497 & 501 \\
\midrule
\textbf{Avg PPA} & \textbf{0.935} & \textbf{0.891} & \textbf{0.925} & \textbf{0.911} & \textbf{0.930} & \textbf{0.880} & \textbf{0.921} & \textbf{0.920} \\
\bottomrule
\end{tabular}%
}
\end{table*}

\noindent \textbf{Task-related QA pairs.}
To evaluate LLM-based radiology agents, we define 11 task types: (a) organ segmentation, (b) anomaly detection, (c) standard end-to-end diagnosis, (d) organ and anomaly joint grounding, (e) diagnosis with grounding clues, (f) organ-wise biomarker calculation, (g) anomaly-wise biomarker calculation, (h) standard report generation, (i) report generation focused on specific biomarkers, (j) report generation focused on both biomarkers and indicators\footnote{An indicator refers to a specific quantification of health status or disorder grading, e.g., CURB-65 or tumor grade.}, and (k) detailed treatment planning. As illustrated in Figure~\ref{fig:EvalPlat}b,d, these tasks are grouped into `Simple', `Moderate', and `Complex' levels according to the required reasoning steps ($<4$, 4--6, and $\geq6$). For example, organ segmentation mainly requires anatomy/modality classification followed by segmentation, whereas treatment planning involves diagnosis, quantification, and report synthesis. Applying the task templates to 2,200 patient records yields 24,200 QA pairs (2,200 $\times$ 11), which are balanced across task types and manually verified. Per-task definitions, required tool chains, and worked examples are given in Appendix~\ref{sec:task_types}.

\noindent \textbf{Radiology Tool Sets.}
To simulate diverse clinical tool environments, we define 10 high-level radiology tool categories: Anatomy Classifier (AC), Modality Classifier (MC), Organ Segmentor (OS), Anomaly Detector (AD), Imaging Diagnoser (ID), Grounded Diagnoser (GD), Biomarker Quantifier (BQ), Indicator Calculator (IC), Report Generator (RG), and Treatment Planner (TP). Each tool is represented by a ``tool card''~(Figure~\ref{fig:EvalPlat}d) specifying its category, properties, capabilities, required/optional inputs, and performance sampled from clinically plausible ranges, enabling diverse tool variations. We further define four evaluation tool-set conditions: \textbf{``Baseline''} provides a minimal solvable set of 12 tools across 10 categories, with BQ and IC each containing two tools for organ- and anomaly-level assessment; \textbf{``Redundant''} adds irrelevant tools at regular (12--15 tools, at most one extra per category), medium (27--34 tools, 2--3 extras per category), and high (fixed 169 tools) levels to test filtering in noisy settings; \textbf{``Insufficient''} creates unsolvable sets through config1 missing an entire category (14--17 tools), config2 modality/anatomy mismatches (15--17 tools), or config3 missing required capabilities (fixed 18 tools), with increasing recognition difficulty; and \textbf{``Differentiated''} offers overlapping tools (17--18 tools) with distinct performance levels, requiring selection of the most accurate tool. The full definition of each tool category, together with its input/output schema and an example tool card, is provided in Appendix~\ref{sec:tool_categories}.

\begin{figure*}[!t]
    \centering
    \includegraphics[width=1\linewidth]{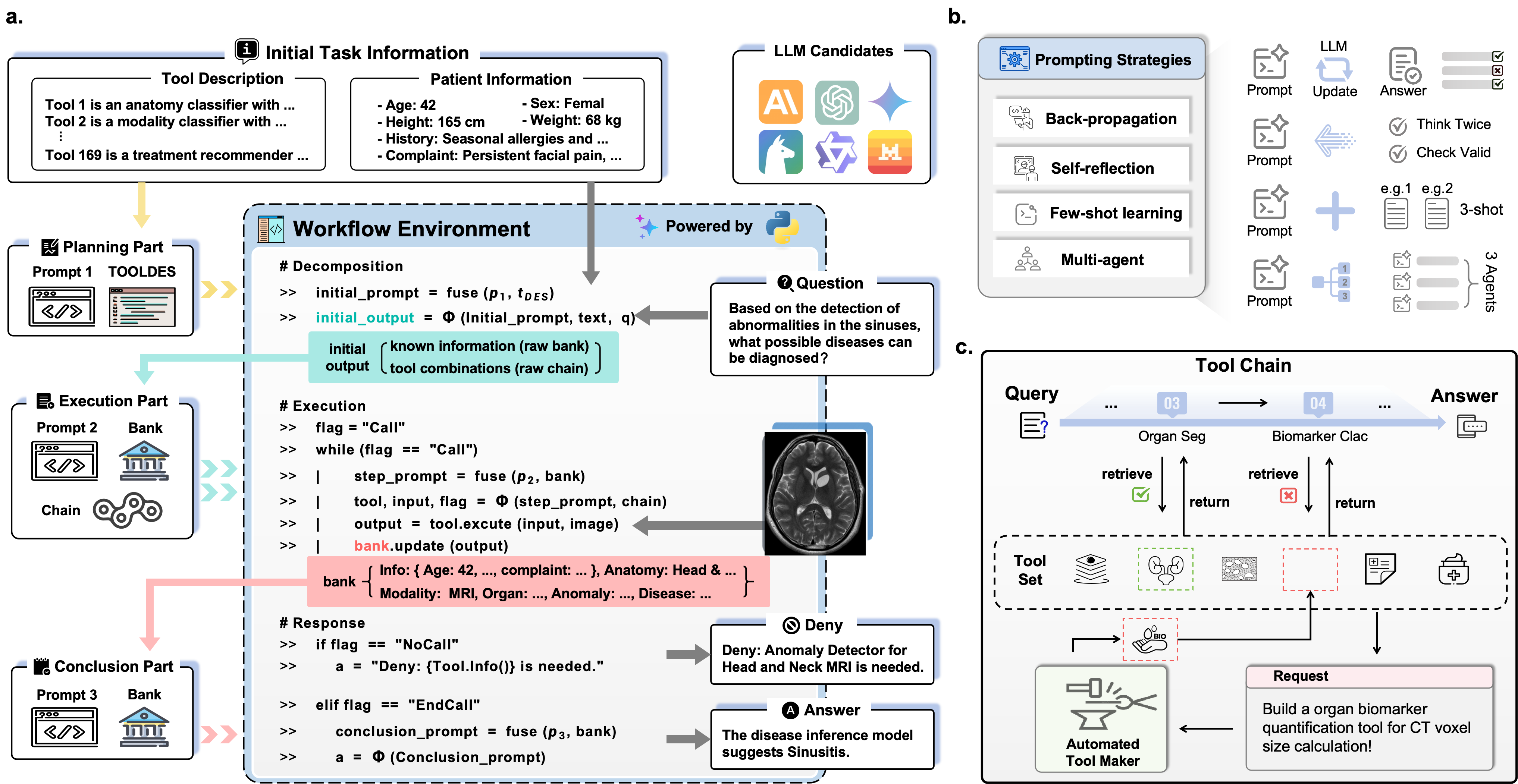}
    \caption{\textbf{Detailed agentic workflow and extended methods.} \textbf{a.} Three-phase architecture consisting of: initialization phase, circulation phase and conclusion phase. \textbf{b.} Four extended prompt engineering methods for performance improvement. \textbf{c.} Layer-specific AutoTB process implemented when no suitable tool is available.}
    \label{fig:Pipeline}
\end{figure*}

\noindent \textbf{Real Radiology Environments.}
In addition to synthetic QA pairs, we construct a test set of 165 real-world cases from Chest Imagenome~\cite{wu2021chest} and RadGenome-BrainMRI datasets~\cite{Lei2025InterpretableBM}, covering three tasks: organ/anomaly grounding, diagnosis, and report generation. Each case preserves the original dataset components, including the clinical question, image, and ground-truth annotation or report. For executable evaluation, we build the real-case tool pool using $\text{M}^3$Builder~\cite{feng2025m}, covering segmentation, detection, and report-generation tools, and further supplement it with nnU-Net~\cite{Isensee2020nnUNetAS} and RP3D-DM~\cite{zheng2024large} for segmentation and diagnosis alternatives. We further instantiate 660 insufficient-tool scenarios from cases in these datasets, enabling evaluation of automated tool building under real conditions. Case-selection criteria and the provenance, modality, and pathology coverage of the two source collections are detailed in Appendix~\ref{sec:real_env_data}, and the construction of the executable tool pool is described in Appendix~\ref{sec:tool_pool}.

\subsection{Benchmarking LLMs as Agent Cores}

\noindent \textbf{Main Workflow Formulation.}
The LLM-based agent system consists of three main components~(Figure~\ref{fig:Pipeline}a): an LLM agent core (\(\Phi\)), a set of specialized tools (\(\mathcal{T}=\{t_1,t_2,\ldots,t_m\}\)), and a memory bank (\(\mathcal{B}\)). Given a clinical query \(\mathcal{Q}\) and patient information, \(\Phi\) first parses the query, decomposes it into sub-tasks according to the available tools in \(\mathcal{T}\), and initializes \(\mathcal{B}\) with relevant patient data. For each sub-task, the agent selects and invokes the appropriate tool, stores intermediate outputs in \(\mathcal{B}\), and uses these outputs to guide subsequent tool calls. Finally, \(\Phi\) synthesizes the accumulated evidence into a response that directly addresses the clinical query.

\noindent \textbf{Prompting and Automated Tool Building.}
To address the limited performance of LLMs on complex tasks, we introduce four prompting enhancement strategies~(Figure~\ref{fig:Pipeline}b). \textbf{Prompt Back-propagation} iteratively refines prompts through three rounds of automated feedback and manual adjustment based on 220 QA pairs under ``Redundant'' and ``Insufficient'' tool-set conditions, after which the refined prompt is applied across the benchmark. \textbf{Self-reflection} asks the agent to verify input correctness before tool invocation, supported by an automated verifier to reduce execution errors. \textbf{Few-shot Learning} provides exemplar workflows and tool templates to standardize planning and tool use, while \textbf{Multi-agent Collaboration} restructures the agent into a planner, executor, and summarizer for task decomposition, I/O management, and final response synthesis. AutoTB follows different protocols in the two benchmark layers. In the synthetic layer, the agent receives M$^3$Builder usage instructions and is evaluated on whether it correctly triggers AutoTB and submits a build request matching the condition-specific missing fields: category for config1, anatomy and modality for config2, and category with the missing capability for config3. No model is trained in this layer; after a request passes verification, the requested tool is simulated as available so that subsequent orchestration can be evaluated. In the real-environment layer, M$^3$Builder~\cite{feng2025m} is actually invoked to train and validate the requested tool, register successful tools in the executable pool, and record construction and usage outcomes in its logs. Appendix~\ref{sec:autotb_protocol} gives the verification rules, executable pseudo-code, and specific traces for both layers.

\noindent \textbf{Real-Case Evaluation Protocol.}
For real-world testing, we use the original clinical question, image, and reference annotation/report from each case without converting them into synthetic patient records. The agent receives the clinical input and must complete the task by selecting from executable radiology tools rather than relying on idealized tool outputs. On the 165-case test set, we evaluate organ/anomaly grounding, diagnosis with grounding clues, and standard report generation using task-specific automatic metrics, including Dice/Jaccard for grounding, Accuracy/Jaccard for diagnosis, and RadGraph-F1/RaTE for reports. We also collect LLM-based ratings and blinded human ratings from two expert radiologists to assess clinical plausibility. To quantify the gap between idealized and executable settings, we define a reported tool-performance ceiling $U_m$ for each automatic metric $m$ by substituting the officially reported performance of the relevant underlying task-specific tool~\cite{feng2025m,Isensee2020nnUNetAS,zheng2024large} into the idealized evaluation protocol. This quantity assumes correct orchestration and reported tool performance, so it is a reference ceiling rather than a theoretical or clinical upper bound. Let $C_m$ be the column-wise best observed score across evaluated systems (not a single deployable system); we compute the relative gap as $\Delta_m=(U_m-C_m)/C_m\times100\%$. For AutoTB, we construct 660 real insufficient-tool scenarios by hiding required tool categories, modality/anatomy-specific capabilities, or functional requirements. The agent must recognize the missing capability, issue a grounded build request, and use the resulting executable tool to complete the augmented workflow; M$^3$Builder logs are used to distinguish construction failures from incorrect tool usage. Formal definitions of all automatic and agent-level metrics are given in Appendix~\ref{sec:eval_metrics}.
\section{Results}
\vspace{-2pt}
\subsection{General Observations}

\begin{table}[!h]
    \centering
    \caption{Task completion rates of recent LLMs across task difficulty levels. }
    \label{tab:current_model_completion}
    \setlength{\tabcolsep}{6pt}
    \small
    \begin{tabular}{lccc}
    \toprule
    \textbf{Model} & \textbf{Regular} & \textbf{Medium} & \textbf{High} \\
    \midrule
    DeepSeek-R1          & 0.600 & \textbf{0.527} & 0.273 \\
    Qwen3-235B           & 0.491 & 0.382 & 0.182 \\
    Claude-Sonnet-4.5    & \textbf{0.691} & 0.509 & \textbf{0.327} \\
    Gemini-3.0-Pro       & 0.572 & 0.484 & 0.260 \\
    GPT-5.2-Chat         & 0.618 & 0.473 & 0.309 \\
    \bottomrule
    \end{tabular}
\end{table}

The evaluated model versions follow their technical reports or model cards~\cite{guo2025deepseek,yang2025qwen3,anthropic2025sonnet45,google2025gemini3pro,openai2025gpt52}.

As shown in Table~\ref{tab:current_model_completion}, recent LLMs still show a clear decline in task completion as difficulty increases; for example, Claude-Sonnet-4.5 drops from 0.691 on regular tasks to 0.509 on medium tasks and 0.327 on high-difficulty tasks. Because large-scale repeated evaluation is costly and model versions evolve rapidly, some analyses below use earlier representative LLMs to diagnose general failure mechanisms rather than rank specific releases, while newer models are evaluated where needed and further tested on real-world clinical data. Overall, the main bottleneck lies in translating tool-level reasoning into reliable end-to-end execution: agents can often form plausible plans, select competitive tools, and complete intermediate milestones, but errors in execution, I/O passing, target alignment, and failure-cause grounding prevent these partial successes from consistently becoming completed tasks.

\begin{figure*}[!h]
    \centering
    \includegraphics[width=1\linewidth]{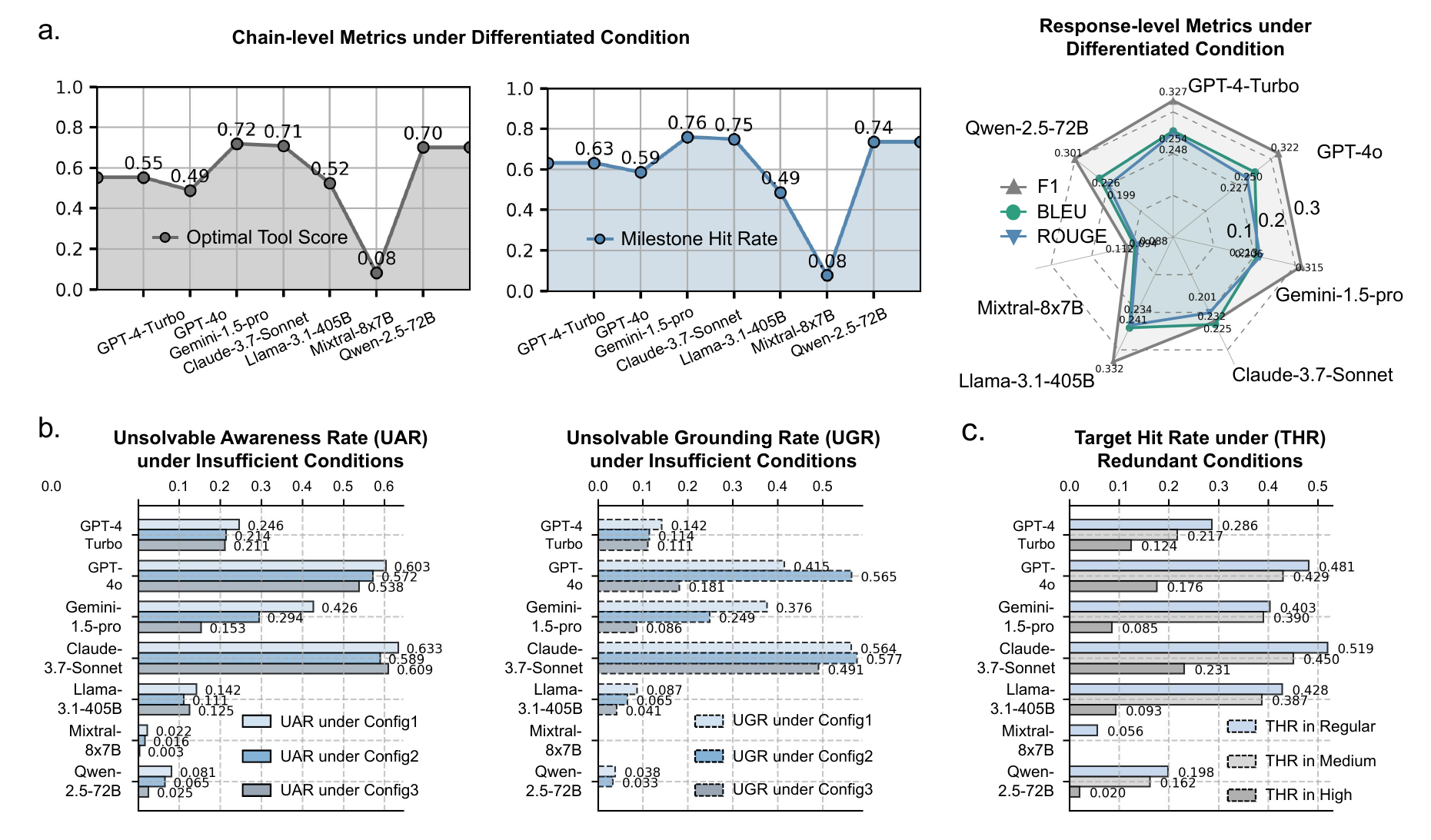}
    \caption{\textbf{Bottleneck analysis under differentiated, insufficient, and redundant tool settings.} \textbf{a.} Tool- and response-level metrics under the ``Differentiated'' setting, showing tool selection, milestone completion, target alignment, and final response synthesis. \textbf{b.} Performance results under “Insufficient” conditions across LLMs. \textbf{c.} Performance results under “Redundant” conditions for leading LLMs.}
    \label{fig:benchmark2}
\end{figure*}

\vspace{3pt}
\noindent \textbf{Observation 1: The Main Bottleneck Lies in Execution Fidelity Rather than Initial Planning}

To separate initial planning from downstream execution, we evaluate tool-chain generation using Levenshtein Distance (LD), False Discovery Rate (FDR), and Tool Matching Accuracy (TMA), and compare them with Execution Completion Rate (ECR) and Pre-Failure Success Percentage (PFSP). As shown in Table~\ref{tab:redundant_metrics}, Claude-Sonnet-4.5 maintains relatively small chain distances as difficulty grows (Exec. Dist.: 0.97 to 1.28; Cons. Dist.: 0.61 to 0.90), and TMA remains 0.617 even under high redundancy. However, ECR and PFSP fall from 0.856/0.822 to 0.425/0.318, indicating that plausible plans increasingly drift during execution through incorrect intermediate outputs or mismatched downstream tool inputs. This gap points to failures in long-chain execution consistency, especially state tracking and intermediate validation.

\begin{table}[!h]
\centering
\caption{Claude-Sonnet-4.5 tool-chain metrics under redundant tool settings. ($\downarrow$: lower is better; $\uparrow$: higher is better).}
\label{tab:redundant_metrics}
\footnotesize
\setlength{\tabcolsep}{6pt}
\begin{tabular}{@{}lccc@{}}
\toprule
\textbf{Metric} & \textbf{Regular} & \textbf{Medium} & \textbf{High} \\
\midrule
Exec. Dist. $\downarrow$ & 0.97 & 1.10 & 1.28 \\
Cons. Dist. $\downarrow$ & 0.61 & 0.73 & 0.90 \\
FD Rate $\downarrow$ & 0.094 & 0.139 & 0.164 \\
TM Acc. $\uparrow$ & 0.815 & 0.704 & 0.617 \\
EC Rate $\uparrow$ & 0.856 & 0.766 & 0.425 \\
PFS \% $\uparrow$ & 0.822 & 0.550 & 0.318 \\
\bottomrule
\end{tabular}
\end{table}

\noindent \textbf{Observation 2: Tool-Level Competence Does Not Reliably Translate into Final Success}

Under the ``Differentiated'' setting, the Optimal Tool Score (OTS, how closely selected tools rank relative to available alternatives) and Milestone Hit Rate (MHR, success in producing key intermediate outputs) show that agents can often choose competitive tools and complete important intermediate steps. Nevertheless, response-level metrics remain much lower: Target Hit Rate (THR, whether the final tool aligns with the required answer format) is limited, and BLEU, ROUGE, and F1 rarely exceed 0.35 (Figure~\ref{fig:benchmark2}a). This gap indicates that intermediate tool success is not sufficient for task completion. The downstream bottleneck lies in target grounding and response synthesis: agents must determine which intermediate result should answer the query, align the final output with the required response format, and integrate tool outputs into a coherent radiological response.

\vspace{3pt}
\noindent \textbf{Observation 3: Tool-Set Perturbations Expose Dependency and Grounding Bottlenecks}

The ``Redundant'' and ``Insufficient'' settings reveal two complementary failure modes. With redundant tools, the challenge shifts from choosing one useful tool to maintaining correct data dependencies among many similar candidates; the ECR and PFS declines in Table~\ref{tab:redundant_metrics} reveal I/O coupling errors and downstream error propagation. With insufficient tools, agents show stronger Unsolvability Awareness Rate (recognizing that a task is impossible) than Unsolvability Grounding Rate (identifying what is missing): they may detect that a task cannot be solved, but struggle to specify whether the missing component is a tool category, a modality-region capability, or a minimum functional requirement (Figure~\ref{fig:benchmark2}b). Together, these perturbation tests show that robust radiology agents require not only stronger LLMs, but also explicit mechanisms for state tracking, dependency validation, target grounding, and failure-cause attribution.

\vspace{-2pt}
\subsection{Performance Enhancement}

\noindent \textbf{Prompting Strategies Mainly Improve Long-Chain Coordination} 

\begin{figure}[!h]
    \centering
    \includegraphics[width=1\linewidth]{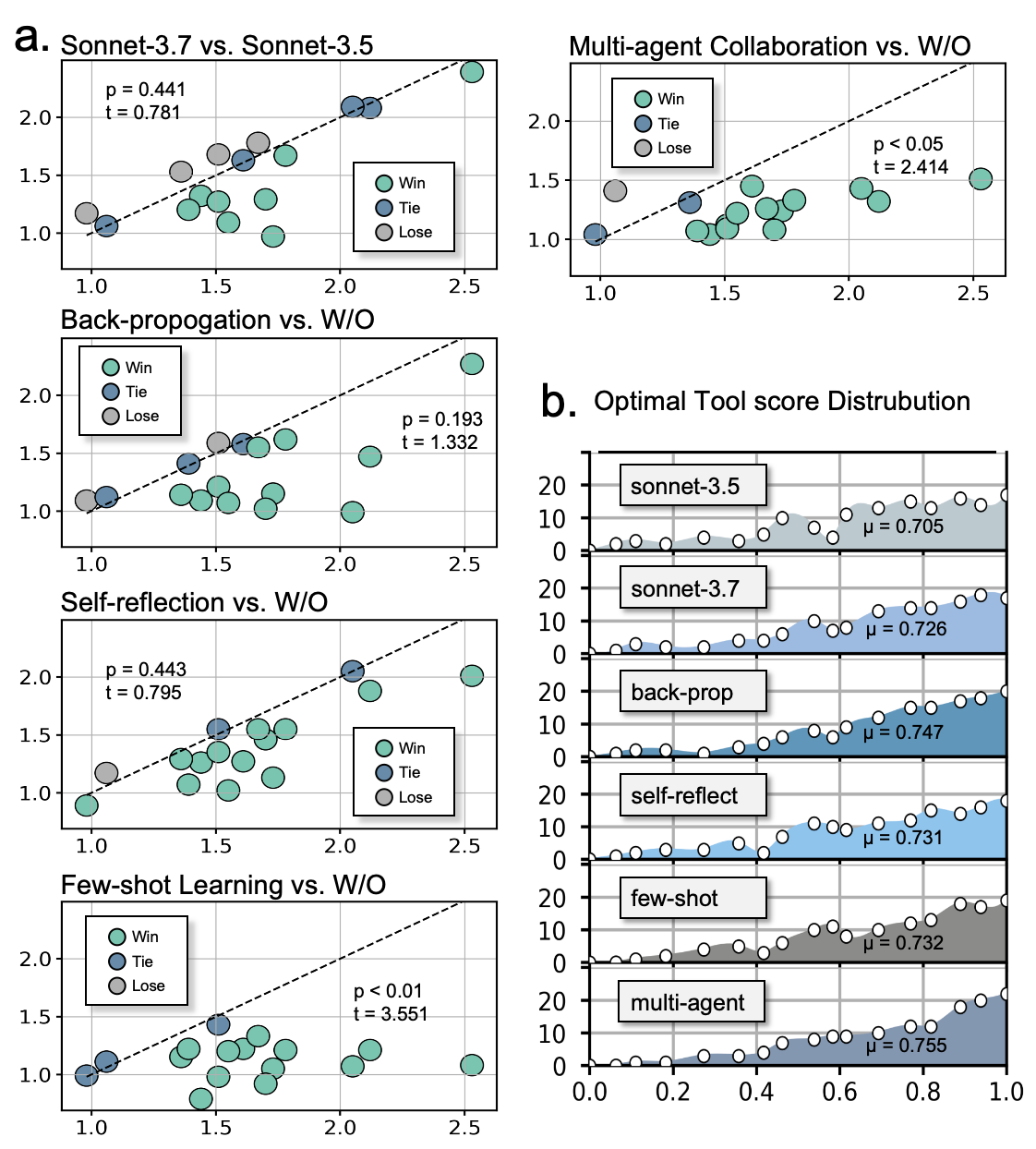}
    \caption{\textbf{Performance improvement comparison among prompt engineering strategies.} a. Significance comparison between strategy groups, where ``W/O'' represents the baseline without prompt engineering, and ``Win'' indicates that the method before ``vs.'' outperforms the latter. b. Performance distribution across six prompting strategies under ``Differentiated'' condition.}
    \label{fig:PromptAblation1}
\end{figure}

Prompting strategies improve agent performance by targeting the coordination bottlenecks identified above rather than simply changing the base model. The integrated strategy increases task completion by 0.095 and 0.058 on simple and moderate tasks, and by up to 48.2\% on complex tasks (Figure~\ref{fig:PromptAblation1}). Among individual strategies, few-shot learning and multi-agent collaboration yield significant wins ($p < 0.05$), whereas self-reflection and back-propagation alone are less stable. Multi-agent collaboration provides the most consistent gains by improving long-chain monitoring and failure checking, raising unsolvability awareness/grounding by about 5\%; back-propagation contributes smaller but useful downstream correction gains (about 3\%) and, together with multi-agent collaboration, yields the largest OTS improvements in the ``Differentiated'' setting (0.021 and 0.029; Figure~\ref{fig:PromptAblation1}b).

\vspace{3pt}
\noindent \textbf{Synthetic AutoTB Request Success Declines with Task Complexity} 

We evaluate the decision-level AutoTB protocol under the synthetic \textit{Insufficient} setting, where required tool categories, modality-specific tools, or capabilities are absent. No model is trained in this layer. A case is counted as a successful AutoTB request only when the agent triggers the mechanism and correctly grounds the fields required by that condition: the missing category in config1, the anatomy--modality pair in config2, or the deficient category and capability in config3. After verification, the requested tool is simulated as available so that the remaining tool-chain orchestration can be evaluated. Across 220 QA pairs, the verified AutoTB success rates are 65.4\% for routine tasks, 44.4\% for advanced tasks, and 26.2\% for complex tasks. The main bottleneck is missing-capability grounding: especially under \textit{Insufficient config3}, the agent may request an inappropriate tool despite recognizing that the original tool set is inadequate. These results therefore characterize build-request grounding and subsequent idealized orchestration, rather than the performance or safety of a newly trained medical tool.

\subsection{Real Case Evaluation}

To evaluate real-world task completion, we benchmark representative proprietary, open-source, medical LLM, medical vision-language, and agentic systems on 165 cases across organ/anomaly grounding, diagnosis with grounding clues, and standard report generation (55 cases each). Unless otherwise specified, agentic experiments use Claude-Sonnet-4.5 as the agent core and an executable M$^3$Builder tool pool covering segmentation, detection, and report generation. We report task-specific automatic metrics, LLM-based ratings, and human ratings from two expert radiologists; AutoTB is separately analyzed under insufficient-tool conditions below.

\begin{table*}[!t]
\centering
\caption{Real-world evaluation across 165 cases from organ/anomaly grounding, diagnosis with grounding clues, and standard report generation. All metrics are higher-is-better. See Appendix~\ref{sec:eval_metrics} for each metric in detail.}
\label{tab:real_case_evaluation}
\scriptsize
\setlength{\tabcolsep}{2pt}
\renewcommand{\arraystretch}{1.12}
\resizebox{\textwidth}{!}{%
\begin{tabular}{@{}l*{12}{c}@{}}
\toprule
\multirow{2}{*}{\textbf{Method}} &
\multicolumn{4}{c}{\begin{tabular}[c]{@{}c@{}}\textbf{Organ and Anomaly Joint Grounding}\\\textbf{(55 cases from Chest Imagenome)}\end{tabular}} &
\multicolumn{4}{c}{\begin{tabular}[c]{@{}c@{}}\textbf{Diagnosis with Grounding Clues}\\\textbf{(55 cases from RadGenome-MRI)}\end{tabular}} &
\multicolumn{4}{c}{\begin{tabular}[c]{@{}c@{}}\textbf{Standard Report Generation}\\\textbf{(55 cases from RadGenome-MRI)}\end{tabular}} \\
\cmidrule(lr){2-5}\cmidrule(lr){6-9}\cmidrule(lr){10-13}
& \textbf{Dice $\uparrow$} & \begin{tabular}[c]{@{}c@{}}\textbf{Jaccard}\\\textbf{Index $\uparrow$}\end{tabular} & \begin{tabular}[c]{@{}c@{}}\textbf{LLM}\\\textbf{Rating $\uparrow$}\end{tabular} & \begin{tabular}[c]{@{}c@{}}\textbf{Human}\\\textbf{Rating $\uparrow$}\end{tabular}
& \textbf{Accuracy $\uparrow$} & \begin{tabular}[c]{@{}c@{}}\textbf{Jaccard}\\\textbf{Index $\uparrow$}\end{tabular} & \begin{tabular}[c]{@{}c@{}}\textbf{LLM}\\\textbf{Rating $\uparrow$}\end{tabular} & \begin{tabular}[c]{@{}c@{}}\textbf{Human}\\\textbf{Rating $\uparrow$}\end{tabular}
& \begin{tabular}[c]{@{}c@{}}\textbf{RadGraph}\\\textbf{F1 $\uparrow$}\end{tabular} & \begin{tabular}[c]{@{}c@{}}\textbf{RaTE}\\\textbf{Score $\uparrow$}\end{tabular} & \begin{tabular}[c]{@{}c@{}}\textbf{LLM}\\\textbf{Rating $\uparrow$}\end{tabular} & \begin{tabular}[c]{@{}c@{}}\textbf{Human}\\\textbf{Rating $\uparrow$}\end{tabular} \\
\midrule
\multicolumn{13}{@{}l}{\textbf{General Closed-source}} \\
GPT-4o~\cite{openai2024gpt4o} & 0.467 & 0.305 & 2.455 & 2.127 & 0.418 & 0.305 & 2.727 & 2.818 & 0.103 & 0.328 & 2.927 & 2.382 \\
o3~\cite{openai2025o3} & 0.479 & 0.315 & 2.636 & 2.455 & 0.436 & 0.335 & 2.764 & 2.873 & 0.115 & 0.345 & 2.964 & 2.455 \\
Gemini-3.0-Pro~\cite{google2025gemini3pro} & 0.489 & 0.324 & 2.909 & 2.727 & 0.455 & 0.348 & 2.818 & 2.909 & 0.125 & 0.360 & 3.000 & 2.509 \\
Claude-Sonnet-4.5~\cite{anthropic2025sonnet45} & 0.503 & 0.336 & \textbf{3.218} & \textbf{3.164} & \textbf{0.476} & \textbf{0.364} & 2.855 & 2.945 & 0.137 & \textbf{0.385} & 3.036 & 2.564 \\
\midrule
\multicolumn{13}{@{}l}{\textbf{General Open-source}} \\
DeepSeek-R1~\cite{guo2025deepseek} & 0.486 & 0.321 & 2.582 & 2.455 & 0.394 & 0.313 & 2.618 & 2.363 & 0.100 & 0.314 & 2.727 & \textbf{2.709} \\
Llama-3.1-405B~\cite{grattafiori2024llama3} & 0.252 & 0.144 & 1.764 & 1.800 & 0.307 & 0.129 & 1.945 & 1.836 & 0.043 & 0.237 & 2.109 & 1.873 \\
Qwen3-235B~\cite{yang2025qwen3} & 0.462 & 0.300 & 2.455 & 2.182 & 0.364 & 0.285 & 2.545 & 2.273 & 0.091 & 0.300 & 2.636 & 2.545 \\
\midrule
\multicolumn{13}{@{}l}{\textbf{Medical LLM}} \\
Meditron3-70B~\cite{sallinen2025llama} & 0.392 & 0.244 & 2.164 & 1.927 & 0.277 & 0.204 & 1.964 & 1.818 & 0.069 & 0.248 & 2.164 & 2.036 \\
MedGemma-27B~\cite{Sellergren2025MedGemmaTR} & 0.403 & 0.252 & 2.218 & 2.000 & 0.291 & 0.212 & 2.036 & 1.909 & 0.075 & 0.255 & 2.218 & 2.091 \\
Baichuan-M3-235B~\cite{Dou2026BaichuanM3MC} & 0.413 & 0.260 & 2.273 & 2.073 & 0.309 & 0.225 & 2.109 & 1.964 & 0.082 & 0.265 & 2.291 & 2.182 \\
\midrule
\multicolumn{13}{@{}l}{\textbf{End-to-end Medical Vision-language Model}} \\
Llava-Med~\cite{li2023llavamed} & 0.110 & 0.058 & 1.636 & 1.800 & 0.295 & 0.089 & 2.036 & 1.818 & 0.007 & 0.138 & 1.533 & 1.845 \\
RadFM~\cite{wu2023towards} & 0.122 & 0.065 & 1.709 & 1.855 & 0.309 & 0.105 & 2.091 & 1.909 & 0.015 & 0.150 & 1.600 & 1.909 \\
MedDr~\cite{He2024GSCoTG} & 0.140 & 0.075 & 1.782 & 1.909 & 0.327 & 0.125 & 2.145 & 2.000 & 0.025 & 0.165 & 1.727 & 2.000 \\
\midrule
\multicolumn{13}{@{}l}{\textbf{Medical Agentic Systems}} \\
MAM~\cite{Zhou2025MAMMM} & 0.406 & 0.255 & 2.273 & 2.055 & 0.351 & 0.219 & 2.345 & 2.218 & 0.077 & 0.262 & 2.800 & 2.418 \\
RadCouncil~\cite{Zeng2024EnhancingLF} & 0.394 & 0.245 & 2.182 & 2.000 & 0.327 & 0.205 & 2.273 & 2.145 & 0.070 & 0.250 & 2.727 & 2.327 \\
AgentClinic~\cite{schmidgall2024agentclinic} & \textbf{0.513} & \textbf{0.345} & 3.091 & 3.091 & 0.455 & 0.355 & \textbf{2.909} & \textbf{3.000} & \textbf{0.145} & 0.370 & \textbf{3.055} & 2.636 \\
\bottomrule
\end{tabular}%
}
\end{table*}

\begin{table}[!t]
\centering
\caption{Comparison between the current per-metric best and the reported tool-performance ceiling. OAG: organ/anomaly grounding; DGC: diagnosis with grounding clues; SRG: standard report generation.}
\label{tab:execution_gap}
\scriptsize
\setlength{\tabcolsep}{3pt}
\begin{tabular}{llccc}
\toprule
\textbf{Task} & \textbf{Metric} & \textbf{Current Best} & \textbf{Ceiling} & \textbf{Gap} \\
\midrule
OAG & Dice & 0.513 & 0.591 & 15.2\% \\
OAG & Jaccard & 0.345 & 0.397 & 15.2\% \\
DGC & Accuracy & 0.476 & 0.577 & 21.2\% \\
DGC & Jaccard & 0.364 & 0.419 & 15.1\% \\
SRG & RadGraph F1 & 0.145 & 0.162 & 11.7\% \\
SRG & RaTE Score & 0.385 & 0.441 & 14.6\% \\
\midrule
\multicolumn{4}{r}{\textbf{Mean gap}} & \textbf{15.5\%} \\
\bottomrule
\end{tabular}
\end{table}

\vspace{3pt}
\noindent \textbf{Real-World Tools Expose a Gap from Reported Tool Ceilings.}

Table~\ref{tab:execution_gap} makes the idealized--executable comparison explicit. The reported tool-performance ceilings are 15.5\% higher on average than the current per-metric best scores. Here, \textit{Current Best} is the column-wise envelope across evaluated systems rather than the output of one deployable agent, and the ceiling transfers officially reported task-specific tool performance into the idealized protocol. It should therefore be interpreted as a computational reference point, not as a theoretical or clinically attainable upper bound. The metric-wise gaps nevertheless consistently show that executable pipelines lose performance beyond idealized tool orchestration, with the largest gap appearing in diagnosis accuracy (21.2\%). This degradation reflects two deployment bottlenecks. First, executable tools have limited and domain-dependent accuracy, as even the strongest grounding and report-generation results remain modest (e.g., best Dice/Jaccard of 0.513/0.345 and best RadGraph-F1 of 0.145). Second, the agent must align tool inputs and outputs, select among models trained on different datasets, and recover from runtime or format mismatches; therefore, a plausible high-level plan does not always lead to a valid final response. Even under these constraints, the agentic systems remain comparable to representative medical agentic baselines: the strongest agentic result reaches 0.455 diagnosis accuracy and 0.145 report RadGraph-F1, close to or above MAM/RadCouncil (0.351/0.327 accuracy and 0.077/0.070 RadGraph-F1), while maintaining similar clinical ratings in diagnosis and report generation. These results suggest that practical gains will depend as much on tool reliability, execution alignment, and failure recovery as on improving the agent core itself.

\begin{table}[!h]
\centering
\caption{Stability analysis on 660 insufficient-tool scenarios.}
\label{tab:autotb_stability}
\scriptsize
\setlength{\tabcolsep}{3pt}
\begin{tabular}{@{}lccc@{}}
\toprule
\textbf{Difficulty} & \begin{tabular}[c]{@{}c@{}}\textbf{Trigger}\\\textbf{Rate $\uparrow$}\end{tabular} & \begin{tabular}[c]{@{}c@{}}\textbf{Exec. Success}\\\textbf{Rate $\uparrow$}\end{tabular} & \begin{tabular}[c]{@{}c@{}}\textbf{Avg. Build}\\\textbf{Time (min) $\downarrow$}\end{tabular} \\
\midrule
Regular (N=220) & 98.2\% & 94.5\% & 12.3 \\
Intermediate (N=220) & 92.7\% & 86.8\% & 18.7 \\
Complex (N=220) & 78.6\% & 65.4\% & 29.4 \\
\bottomrule
\end{tabular}
\end{table}

\begin{table}[!h]
\centering
\caption{Failure mode analysis (N=100).}
\label{tab:autotb_failure_modes}
\scriptsize
\setlength{\tabcolsep}{3pt}
\begin{tabular}{@{}p{0.25\linewidth}cp{0.50\linewidth}@{}}
\toprule
\textbf{Failure Category} & \textbf{Pct. $\downarrow$} & \textbf{Description} \\
\midrule
Usage Error & 48\% & Agent correctly identified need but misconfigured tool parameters \\
Tool Deficiency & 27\% & Available tools lacked required functionality \\
Environment Error & 16\% & External dependencies or API failures \\
Planning Error & 9\% & Agent failed to identify when tool building was necessary \\
\bottomrule
\end{tabular}
\end{table}

\begin{table}[!h]
\centering
\caption{Clinical utility on real-world cases (N=30).}
\label{tab:autotb_clinical_utility}
\scriptsize
\setlength{\tabcolsep}{3pt}
\begin{tabular}{@{}lccc@{}}
\toprule
\textbf{Difficulty} & \begin{tabular}[c]{@{}c@{}}\textbf{AutoTB}\\\textbf{Trigger Rate}\end{tabular} & \begin{tabular}[c]{@{}c@{}}\textbf{Report Quality}\\\textbf{(Human 1--5) $\uparrow$}\end{tabular} & \begin{tabular}[c]{@{}c@{}}\textbf{Clinical Utility}\\\textbf{(Human 1--5) $\uparrow$}\end{tabular} \\
\midrule
Regular (N=10) & 100\% & 3.7 & 4.2 \\
Intermediate (N=10) & 90\% & 3.2 & 3.6 \\
Complex (N=10) & 60\% & 2.4 & 2.1 \\
\bottomrule
\end{tabular}
\end{table}

\vspace{3pt}
\noindent \textbf{AutoTB Benefits Clear Missing-Capability Cases but Remains Execution-Limited.}

We evaluate executable AutoTB in a multi-stage real-environment setting: stability testing across 660 insufficient-tool scenarios, failure categorization on 100 cases, and human clinical-utility assessment on 30 real-world cases. Unlike the synthetic protocol, M$^3$Builder is actually invoked here: it trains and validates the requested model, registers a successful tool in the executable pool, and records build and downstream usage outcomes in its logs. Tables~\ref{tab:autotb_stability}--\ref{tab:autotb_clinical_utility} show that AutoTB substantially improves solvability in regular and intermediate scenarios, where trigger rates remain high (98.2\% and 92.7\%), execution success is strong (94.5\% and 86.8\%), and clinical utility is favorable (4.2 and 3.6). In complex scenarios, however, the trigger rate falls to 78.6\%, execution success drops to 65.4\%, average build time increases to 29.4 minutes, and clinical utility declines to 2.1. Failure analysis further shows that the main bottleneck is operational precision rather than high-level awareness: usage errors dominate (48\%) while planning errors are rare (9\%), indicating that agents often know \emph{when} to trigger AutoTB but struggle to specify precise parameters, interfaces, and configurations for complex tool generation. Thus, AutoTB shifts failures from ``missing capability'' to ``fragile capability construction'', motivating schema-guided tool synthesis, parameter validation, dry-run execution checks, and tighter workflow integration.

\vspace{-2pt}
\section{Related Work}
\noindent \textbf{LLM-based Agents in Medicine.} The evolution of medical AI has transitioned from static classifiers to dynamic LLM-based agents capable of multi-step reasoning. Recent frameworks have adopted collaborative multi-agent strategies~\cite{tang2024medagents} or integrated clinical calculators, retrieved guidelines, and specialized analysis tools~\cite{jin2024agentmd, wang2025medagent} to support more transparent decision making. In the imaging domain, architectures such as ChatCAD~\cite{wang2024interactive} use LLMs to synthesize the outputs of specialized vision models. MedAgentBench~\cite{jiang2025medagentbench}, meanwhile, evaluates agents primarily through interactive EHR tasks; it does not cover the multimodal, tool-integrated radiology workflows here.

\noindent \textbf{Synthetic Data Generation and Validation.} Parallel to agent development, the data scarcity--privacy tension has motivated methods ranging from masking and reconstructing de-identified clinical notes~\cite{ren2025synthetic4health} to prompting LLMs to generate synthetic patient--physician conversations~\cite{haider2025synthetic} and large-scale medical image--text training data~\cite{Yilmaz2025ASD}. Validation has accordingly expanded beyond lexical overlap: \citet{sella2025preserving} introduced the Quality--Privacy Score (QPS) to combine data-quality and privacy measures, while \citet{adams2025fidelity} systematically evaluated fidelity, utility, and re-identification risks. In radiology, \citet{sun2025generative} generated synthetic chest-radiology reports and controlled report errors for training and evaluating error detectors. This line of work remains largely text-centered, leaving structured multimodal synthesis for tool-integrated radiology workflows underexplored.

\vspace{-2pt}
\section{Conclusion}

We introduced \textbf{RadA-BenchPlat}, a two-layer benchmark for evaluating radiology agents across planning, tool selection, execution, and failure awareness under tool conditions. Our results show that the main bottleneck lies in execution fidelity rather than high-level planning, especially in long-chain scenarios. By linking synthetic stress tests with real-world execution, RadA-BenchPlat further shows that practical progress depends on reliable tools, aligned interfaces, and failure recovery in addition to stronger LLM cores. Prompting strategies and AutoTB help narrow this gap, but persistent usage errors suggest that agents often know \emph{when} to act without reliably knowing \emph{how} to use the required tools. Overall, these results define the capability boundaries of radiology agents and expose the gap between idealized orchestration and deployment, motivating broader real-world validation.

\section*{Limitations}

Our study combines synthetic stress testing with real-world execution, but the connection between these two layers remains incomplete. The synthetic layer enables controlled diagnosis of planning, tool selection, and execution failures, yet cannot fully capture institution-specific protocols, tool behavior, data distributions, or workflow constraints in clinical environments. In addition, our real-environment evaluation is based on 165 cases, which limits the statistical power of cross-system comparisons. Future work should narrow this synthetic--real gap through confidence intervals, multi-center validation, and adaptive benchmarks whose simulated tool ecosystems better reflect evolving radiology practice.

\section*{Acknowledgements}

This work is supported by the National Natural Science Foundation of China (No. 24Z031503678). ChatGPT was used to refine the writing of parts of this manuscript and not to generate its content.


\bibliography{reference}

\appendix
\section{Appendix}

\subsection{Terminology}

\begin{figure*}[!t]
    \centering
    \includegraphics[width=1\linewidth]{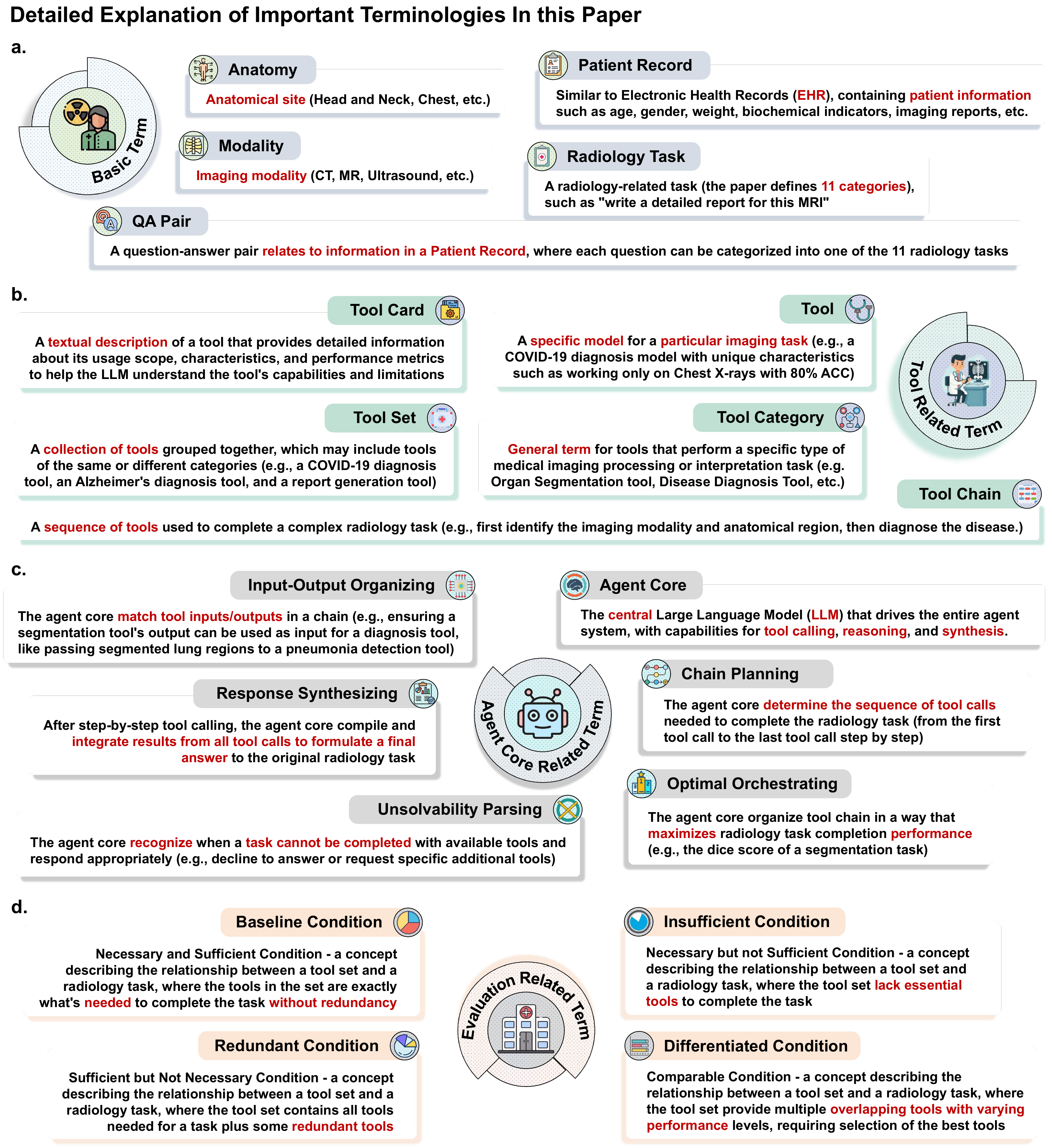}
    \vspace{1mm}
    \caption{\textbf{A detailed explanation of 20 important terminologies used in this paper.} \textbf{a.} Basic terms commonly used throughout the paper. \textbf{b.} Sophisticated definitions of tool-related concepts. \textbf{c.} Agent-related terms including five key abilities. \textbf{d.} Four radiology environments featuring various tool combinations for different evaluation purposes.}
    \label{fig:Terminology}
\end{figure*}

Key terminologies used throughout our work are defined in Extended Figure.~\ref{fig:Terminology}. 

\subsection{Construction Pipeline of the Synthetic Layer}
\label{sec:construction_pipeline}

\begin{figure*}[!t]
    \centering
    \includegraphics[width=1\linewidth]{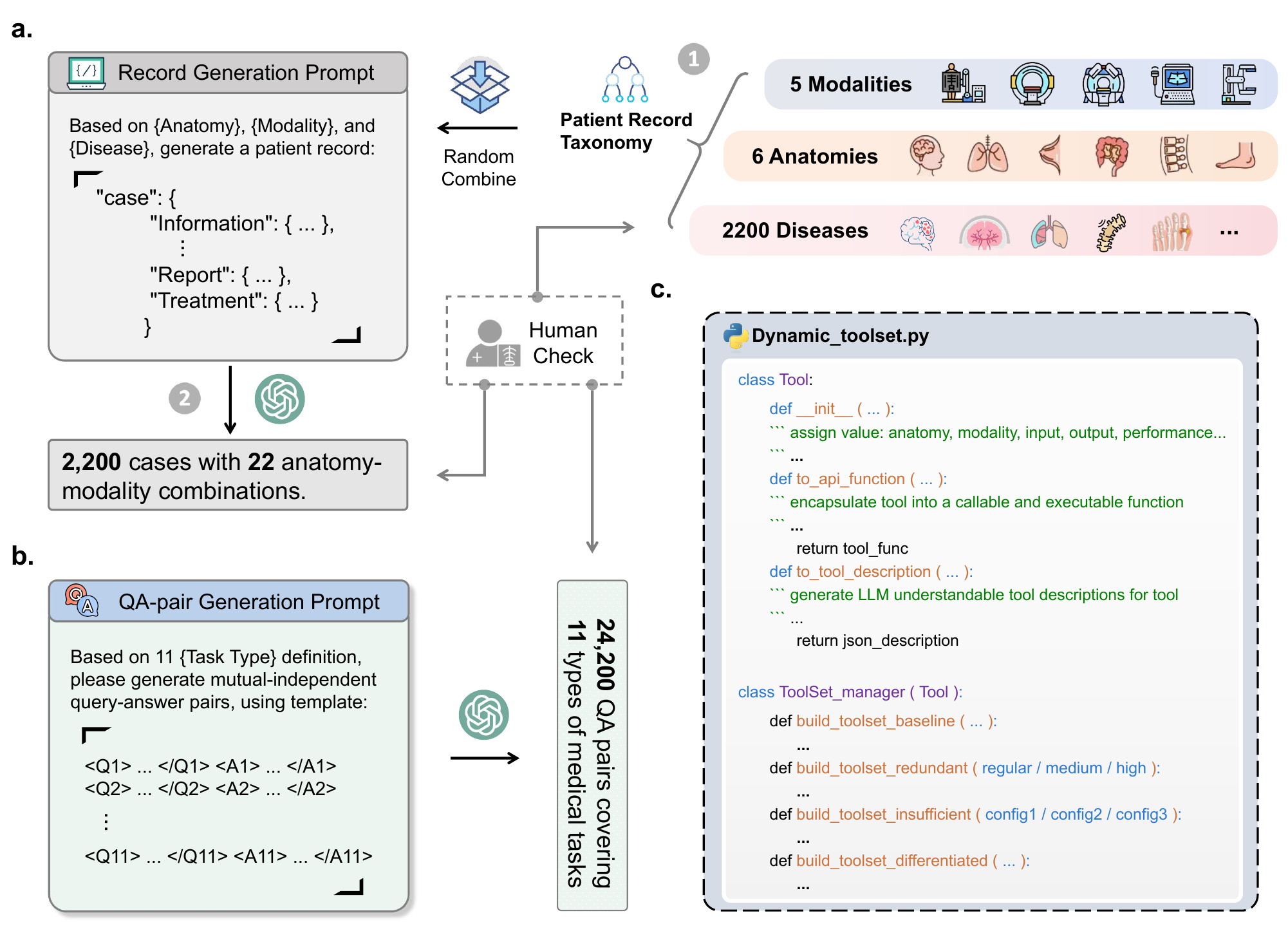}
    \caption{\textbf{Methods for building the synthetic layer of RadA-BenchPlat.} \textbf{a.} Patient record generation based on the modality--anatomy--disease taxonomy, developed through structured templates and followed by a human check. \textbf{b.} QA-pair generation through an ``11 QAs per record'' prompt template powered by GPT-4. \textbf{c.} Dynamic tool-set simulation built on a shared tool-function implementation.}
    \label{fig:PlatMethod}
\end{figure*}

Figure~\ref{fig:PlatMethod} shows how the synthetic layer is instantiated. We describe the generation mechanics here; the resulting taxonomy, task list, and tool-set statistics are reported in the main text.

\vspace{3pt}
\noindent \textbf{Record generation (Figure~\ref{fig:PlatMethod}a).}
An anatomy and a modality are sampled and combined into one of the valid anatomy--modality pairs, and a condition is then drawn from the disease list of that pair. The resulting \{Anatomy\}, \{Modality\}, \{Disease\} triple is substituted into a single record-generation prompt whose output is constrained to a fixed skeleton with named blocks for patient information, imaging findings, report, and treatment. This skeleton is what keeps records field-aligned, so that the same task templates apply to every record and the clinician review can target individual fields instead of free-form text. Records failing the human check are returned for correction rather than discarded.

\vspace{3pt}
\noindent \textbf{QA-pair generation (Figure~\ref{fig:PlatMethod}b).}
Instead of prompting once per task, a single prompt receives one record together with the definitions of all 11 task types and emits the whole set in one pass, following a \texttt{<Q1>...</Q1> <A1>...</A1>} template. The prompt requires the pairs to be mutually independent, so a later question cannot be answered by reusing an earlier answer and each item exercises its intended tool chain in isolation.

\vspace{3pt}
\noindent \textbf{Tool-set simulation (Figure~\ref{fig:PlatMethod}c).}
Tool-set conditions are generated programmatically rather than written per case. A \texttt{Tool} class holds the attributes of one tool card and exposes two views of it: \texttt{to\_api\_function()} wraps the tool as a callable with its declared input/output signature, and \texttt{to\_tool\_description()} renders the LLM-facing description. A \texttt{ToolSet\_manager} then assembles condition-specific pools through \texttt{build\_toolset\_baseline}, \texttt{build\_toolset\_redundant} (regular/medium/high), \texttt{build\_toolset\_insufficient} (config1/config2/config3), and \texttt{build\_toolset\_differentiated}. Since every condition is produced by the same code path from the same tool attributes, what the agent observes differs only in tool availability, and the four conditions can be regenerated deterministically for a new case set.

\subsection{Clinician Validation of the Dataset}
\label{sec:clinician_validation}

The three clinical raters independently assigned a binary pass/fail judgment for each quality criterion. Two raters were from Renmin Hospital of Wuhan University and had more than 10 years of clinical experience, while the third rater was from Shanghai Ninth People’s Hospital and had 8 years of clinical experience. The main-text results aggregate these judgments across raters; this subsection defines the criteria and aggregation procedure.

Specifically, \textbf{Clinical Consist.} evaluates whether the patient record, imaging findings, diagnosis, biomarkers, indicators, report, and treatment description are mutually consistent. \textbf{Info. Validity} checks whether the included clinical attributes are medically plausible and sufficiently complete. \textbf{Tool Necessity} verifies that answering the QA pair requires the intended radiology tools rather than shortcut reasoning alone. \textbf{No Leakage} checks that the question, metadata, or prompt does not directly reveal the answer. \textbf{Reasoning Depth} evaluates whether the item requires the intended level of multi-step clinical and tool-use reasoning. \textbf{Ambiguity} measures whether the task target and reference answer are clear enough to support reliable evaluation. \textbf{Professionalism} assesses the use of appropriate radiology terminology and reporting style. \textbf{Bias/Diversity} checks whether the sampled items avoid obvious demographic or disease shortcuts while maintaining diversity across anatomy, modality, disease, and patient profiles. For each criterion, Avg PPA is computed as $\frac{1}{3}\sum_{r=1}^{3} n_{r}^{+}/550$, where $n_{r}^{+}$ is the number of positive judgments from rater $r$. It provides a criterion-wise estimate of how consistently clinicians accepted the sampled dataset items.

\subsection{Detailed Explanations for Tool Categories}
\label{sec:tool_categories}
We explain the 10 tool categories here: 
\begin{itemize}[itemsep=0pt, parsep=0pt, leftmargin=*]
    \item \textbf{Anatomy Classifier~(AC)} predicts the anatomical region shown in the input image.
    \item \textbf{Modality Classifier~(MC)} predicts the imaging modality of the input image.
    \item \textbf{Organ Segmentor~(OS)} is a tool to predict the dense segmentation masks for a certain organ on the input image.
    \item \textbf{Anomaly Detector~(AD)} is a detection tool to predict the anomaly region masks or box coordinates for a certain anomaly type based on the input image.
    \item \textbf{Imaging Diagnoser~(ID)} is a diagnosis tool to predict related diseases solely based on the input image.
    \item \textbf{Grounded Diagnoser~(GD)} is a diagnosis tool to predict related diseases based on the input image, text information, and grounding clues.
    \item \textbf{Biomarker Quantifier~(BQ)} calculates radiological biomarkers from the input image and organ- or anomaly-level dense masks.
    \item \textbf{Indicator Calculator~(IC)} calculates or predicts clinical indicators (e.g., tumor grade) from the input image and organ- or anomaly-level dense masks.
    \item \textbf{Report Generator~(RG)} is a text generation tool to predict radiology reports.
    \item \textbf{Treatment Planner~(TP)} is a treatment planning tool~(or system) to provide treatment plans based on current clinical findings.
\end{itemize}

\subsection{Detailed Explanations for Task Types}
\label{sec:task_types}
We explain the 11 radiology tasks here with their detailed tool category chains:

\begin{itemize}[itemsep=0pt, parsep=0pt, leftmargin=*]
    \item \textbf{Organ mask annotation} predicts a dense mask for a specified organ. \textbf{Chain:} \textit{anatomy classification → modality classification → organ segmentation}
    \item \textbf{Anomaly region annotation} predicts a regional mask for a specified anomaly type. \textbf{Chain:} \textit{anatomy classification → modality classification → anomaly detection}
    \item \textbf{Organ-wise biomarker calculation} is a task to calculate or measure some organ-wise biomarkers. \textbf{Chain:} \textit{anatomy classification → modality classification → organ segmentation → biomarker quantification}
    \item \textbf{Anomaly-wise biomarker calculation} is a task to calculate or measure some anomaly-wise biomarkers. \textbf{Chain:} \textit{anatomy classification → modality classification → anomaly detection → anomaly quantification}
    \item \textbf{Organ-and-anomaly-wise image interpretation} is a task to interpret the image and answer the user query on organs or anomaly regions or both. \textbf{Chain:} \textit{anatomy classification → modality classification → [organ segmentation, anomaly detection]}
    \item \textbf{Disease diagnosis without visual clues} produces an end-to-end diagnosis from a radiology-centered patient record. \textbf{Chain:} \textit{anatomy classification → modality classification → disease diagnosis}
    \item \textbf{Disease diagnosis with visual clues} produces a diagnosis from the original images together with relevant organ or anomaly segmentation annotations. \textbf{Chain:} \textit{anatomy classification → modality classification → [organ segmentation, anomaly detection] → grounded diagnosis}
    \item \textbf{Common report generation} is a basic radiology routine task to interpret a certain radiology image into free-text descriptions. \textbf{Chain:} \textit{anatomy classification → modality classification → anomaly detection → disease diagnosis → report generation}
    \item \textbf{Report generation focusing on specific biomarkers} emphasizes specified biomarker findings in the final report. \textbf{Chain:} \textit{anatomy classification → modality classification → [organ segmentation, anomaly detection] → [organ biomarker quantification, anomaly biomarker quantification] → report generation}
    \item \textbf{Report generation focusing on specific biomarkers and indicators} emphasizes specified biomarker and indicator findings in the final report. \textbf{Chain:} \textit{anatomy classification → modality classification → [organ segmentation, anomaly detection] → disease diagnosis → [organ biomarker quantification, anomaly quantification] → indicator calculation → report generation}
    \item \textbf{Treatment planning} is a task to provide a treatment plan for a patient. \textbf{Chain:} \textit{anatomy classification → modality classification → [organ segmentation, anomaly detection] → disease diagnosis → [organ biomarker quantification, anomaly quantification] → indicator calculation → report generation → treatment planning}
\end{itemize}

\subsection{Evaluation Metrics}
\label{sec:eval_metrics}

Our evaluation metrics assess the agent core from five complementary capabilities: \textit{chain planning}, \textit{optimal orchestrating}, \textit{I/O organizing}, \textit{response synthesizing}, and \textit{unsolvability parsing}. Unless otherwise specified, higher values indicate better performance for rates, accuracies, and text/clinical-quality scores, while lower values indicate better performance for distances, false discovery rates, build time, and failure percentages. For a QA pair, let $G=(g_1,\ldots,g_L)$ denote the ground-truth tool-category chain, $P=(p_1,\ldots,p_M)$ the planned chain, $E=(e_1,\ldots,e_K)$ the executed chain, and $N$ the number of evaluated QA pairs. We use $\mathbf{1}[\cdot]$ as the indicator function.

\vspace{3pt}
\noindent \textbf{Metrics on Chain Planning and Execution Consistency.}
These metrics compare tool-category-wise chains generated or executed by the agent with the ground-truth task decomposition.
\begin{itemize}[itemsep=0.1cm, parsep=0pt, leftmargin=*]
    \item \textbf{Levenshtein Distance~(LD)} is the minimum number of insertion, deletion, and substitution operations needed to transform one chain into another:
    \vspace{-2pt}
    \[
        \mathrm{LD}(X,G)=\min_{\pi:X\rightarrow G}\sum_{o\in\pi} c(o), \quad c(o)=1,
    \]
    where $X$ can be $P$ or $E$. We report \textbf{Plan Dist.} as $\mathrm{LD}(P,G)$, \textbf{Exec. Dist.} as $\mathrm{LD}(E,G)$, and \textbf{Cons. Dist.} as $\mathrm{LD}(P,E)$. Lower values indicate closer chain matches.
    \item \textbf{False Discovery Rate~(FDR)} measures the proportion of executed or planned tools that are not required by the ground-truth chain:
    \vspace{-2pt}
    \[
        \mathrm{FDR}(X,G)=\frac{|\{x_i\in X:x_i\notin G\}|}{\max(|X|,1)}.
    \]
    It penalizes unnecessary or redundant tool usage, especially under redundant tool-set conditions.
    \item \textbf{Tool Matching Accuracy~(TMA)} measures order-sensitive matching between a predicted chain and the ground-truth chain:
    \vspace{-2pt}
    \[
        \mathrm{TMA}(X,G)=\frac{1}{L}\sum_{i=1}^{L}\mathbf{1}[i\le |X|\ \mathrm{and}\ x_i=g_i].
    \]
    It reflects whether the same tool category appears at the correct step.
\end{itemize}

\vspace{3pt}
\noindent \textbf{Metrics on Optimal Orchestrating.}
These metrics evaluate whether the agent can choose the most suitable concrete tool within a required tool category.
\begin{itemize}[itemsep=0.1cm, parsep=0pt, leftmargin=*]
    \item \textbf{Optimal Tool Score~(OTS)} evaluates tool-selection quality when multiple tools can satisfy the same category demand. For step $t$, let $N_t$ be the number of suitable tools and $R_t$ the performance rank of the selected tool, with $R_t=1$ for the best tool:
    \vspace{-2pt}
    \[
        \begin{aligned}
        \mathrm{OTS}_t &= \frac{N_t-R_t+1}{N_t},\\
        \mathrm{OTS} &= \frac{1}{T}\sum_{t=1}^{T}\mathrm{OTS}_t.
        \end{aligned}
    \]
    The best available tool receives 1.0, and lower-ranked tools receive smaller scores.
\end{itemize}

\vspace{3pt}
\noindent \textbf{Metrics on I/O Organizing.}
These metrics measure whether the agent can correctly manage tool inputs, intermediate outputs, and downstream dependencies during multi-step execution. For case $j$, let $L_j$ be the ground-truth chain length and $s_j$ the number of successfully executed steps before termination or failure.
\begin{itemize}[itemsep=0.1cm, parsep=0pt, leftmargin=*]
    \item \textbf{Execution Completion Rate~(ECR)} measures the proportion of cases in which the whole required chain is executed without input/output, format, or runtime failures:
    \vspace{-2pt}
    \[
        \mathrm{ECR}=\frac{1}{N}\sum_{j=1}^{N}\mathbf{1}[s_j=L_j].
    \]
    \item \textbf{Pre-Failure Success Percentage~(PFSP)} measures partial execution progress among failed cases. Let $\mathcal{F}=\{j:s_j<L_j\}$ be the failed-case set:
    \vspace{-2pt}
    \[
        \mathrm{PFSP}=\frac{1}{|\mathcal{F}|}\sum_{j\in\mathcal{F}}\frac{s_j}{L_j}.
    \]
    This metric distinguishes early failures from failures occurring after several successful tool calls.
\end{itemize}

\vspace{3pt}
\noindent \textbf{Metrics on Response Synthesizing.}
These metrics evaluate whether the agent converts tool outputs into a useful final answer and whether the response matches expert-provided references.
\begin{itemize}[itemsep=0.1cm, parsep=0pt, leftmargin=*]
    \item \textbf{Task Completion Rate~(TCR)} measures end-to-end task success. Let $c_j=1$ if case $j$ completes the required workflow and satisfies the task target, otherwise $c_j=0$:
    \vspace{-2pt}
    \[
        \mathrm{TCR}=\frac{1}{N}\sum_{j=1}^{N}c_j.
    \]
    \item \textbf{Target Hit Rate~(THR)} measures whether the final output type or final tool matches the required response target:
    \vspace{-2pt}
    \[
        \mathrm{THR}=\frac{1}{N}\sum_{j=1}^{N}\mathbf{1}[\hat{y}^{\mathrm{type}}_j=y^{\mathrm{type}}_j].
    \]
    \item \textbf{Milestone Hit Rate~(MHR)} measures whether a task-specific key intermediate result is produced. Let $m_j=1$ if the milestone for case $j$ is achieved, otherwise $m_j=0$:
    \vspace{-2pt}
    \[
        \mathrm{MHR}=\frac{1}{N}\sum_{j=1}^{N}m_j.
    \]
    Milestones include outputs such as organ masks, anomaly localization, biomarker values, preliminary diagnoses, or report components.
    \item \textbf{BLEU}, \textbf{ROUGE}, and \textbf{F1} measure lexical overlap between generated text $Y$ and reference text $Y^*$. BLEU is computed from clipped $n$-gram precision $p_n$ and brevity penalty $\mathrm{BP}$:
    \vspace{-2pt}
    \[
        \mathrm{BLEU}=\mathrm{BP}\cdot\exp\left(\sum_{n=1}^{N_g}w_n\log p_n\right).
    \]
    ROUGE recall and word-overlap F1 are summarized as:
    \vspace{-2pt}
    \[
        \begin{aligned}
        \mathrm{ROUGE} &= \frac{|\mathrm{grams}(Y)\cap\mathrm{grams}(Y^*)|}{|\mathrm{grams}(Y^*)|},\\
        \mathrm{F1} &= \frac{2\cdot \mathrm{Precision}\cdot \mathrm{Recall}}{\mathrm{Precision}+\mathrm{Recall}}.
        \end{aligned}
    \]
\end{itemize}

\vspace{3pt}
\noindent \textbf{Metrics on Unsolvability Parsing.}
These metrics focus on insufficient-tool conditions, where a task cannot be solved because a required category, anatomy--modality capability, or functional requirement is missing.
\begin{itemize}[itemsep=0.1cm, parsep=0pt, leftmargin=*]
    \item \textbf{Unsolvability Awareness Rate~(UAR)} measures whether the agent recognizes that the current tool set cannot solve the task. Let $a_j=1$ if the agent correctly refuses an unsolvable task, otherwise $a_j=0$:
    \vspace{-2pt}
    \[
        \mathrm{UAR}=\frac{1}{N}\sum_{j=1}^{N}a_j.
    \]
    \item \textbf{Unsolvability Grounding Rate~(UGR)} measures whether the agent identifies the specific reason for unsolvability. Let $u_j=1$ if the missing category, capability, or requirement is correctly specified, otherwise $u_j=0$:
    \vspace{-2pt}
    \[
        \mathrm{UGR}=\frac{1}{N}\sum_{j=1}^{N}u_j.
    \]
\end{itemize}

\vspace{3pt}
\noindent \textbf{Metrics for Real-Case and AutoTB Evaluation.}
For real-world executable evaluation, we additionally use task-specific clinical metrics and tool-building metrics.
\begin{itemize}[itemsep=0.1cm, parsep=0pt, leftmargin=*]
    \item \textbf{Dice} and \textbf{Jaccard Index} evaluate organ/anomaly grounding by measuring spatial overlap between predicted mask $M$ and reference mask $M^*$:
    \vspace{-2pt}
    \[
        \begin{aligned}
        \mathrm{Dice} &= \frac{2|M\cap M^*|}{|M|+|M^*|},\\
        \mathrm{Jaccard} &= \frac{|M\cap M^*|}{|M\cup M^*|}.
        \end{aligned}
    \]
    \item \textbf{Accuracy} and \textbf{Jaccard Index} evaluate diagnosis with grounding clues. For predicted diagnosis set $D$ and reference set $D^*$:
    \vspace{-2pt}
    \[
        \begin{aligned}
        \mathrm{Accuracy} &= \frac{1}{N}\sum_{j=1}^{N}\mathbf{1}[\hat{d}_j=d_j],\\
        \mathrm{Jaccard}(D,D^*) &= \frac{|D\cap D^*|}{|D\cup D^*|}.
        \end{aligned}
    \]
    \item \textbf{RadGraph-F1} and \textbf{RaTE Score} evaluate report generation. RadGraph-F1 follows the standard F1 form over extracted clinical entities and relations, while RaTE measures radiology-oriented textual alignment with the reference report.
    \item \textbf{LLM Rating} is an automatic clinical-quality score assigned by three independent LLM evaluators: GPT-5.2-Chat~\cite{openai2025gpt52}, Gemini-3.0-Pro~\cite{google2025gemini3pro}, and DeepSeek-R1~\cite{guo2025deepseek}. For each real-world case, the evaluator receives the clinical task, generated answer, and corresponding reference information, and scores whether the answer is clinically plausible, complete, and aligned with the task target. The final LLM Rating is the average score across cases and evaluator models:
    \vspace{-2pt}
    \[
        \mathrm{LLMRating}=\frac{1}{3N}\sum_{j=1}^{N}\sum_{m\in\mathcal{M}} r_{j,m}^{\mathrm{LLM}}.
    \]
    Here, $\mathcal{M}$ contains GPT-5.2-Chat, Gemini-3.0-Pro, and DeepSeek-R1, and $r_{j,m}^{\mathrm{LLM}}$ is the score assigned by evaluator model $m$ for case $j$. Averaging across the three evaluators reduces dependence on a single model's preference or scoring bias.
    \item \textbf{Human Rating} is the average clinical-quality score assigned by expert radiologists under blinded review:
    \vspace{-2pt}
    \[
        \mathrm{HumanRating}=\frac{1}{2N}\sum_{j=1}^{N}\sum_{h=1}^{2} r_{j,h}^{\mathrm{Human}},
    \]
    where $r_{j,h}^{\mathrm{Human}}$ is the score assigned to case $j$ by radiologist $h$.
    \item \textbf{AutoTB Trigger Rate}, \textbf{Execution Success Rate}, and \textbf{Average Build Time} evaluate automated tool building. Let $b_j=1$ if AutoTB is triggered and $q_j=1$ if the augmented workflow succeeds:
    \vspace{-2pt}
    \[
        \begin{aligned}
        \mathrm{TriggerRate} &= \frac{1}{N}\sum_{j=1}^{N}b_j,\\
        \mathrm{ExecSuccess} &= \frac{1}{N}\sum_{j=1}^{N}q_j,\\
        \mathrm{AvgBuildTime} &= \frac{1}{N_b}\sum_{j=1}^{N_b} t_j.
        \end{aligned}
    \]
    Here, $N_b$ is the number of tool-building attempts and $t_j$ is the build time for attempt $j$.
\end{itemize}

\subsection{Real-Environment Data Collection}
\label{sec:real_env_data}

This subsection details how we construct the 165 real-world cases used in the real-environment layer. Unlike the synthetic layer, these cases are not converted into synthetic patient records. Instead, each case keeps its original clinical input, imaging data, and reference annotation or report so that the agent must interact with executable tools under realistic data constraints. The final real-case set contains 55 organ/anomaly grounding cases from Chest Imagenome and 110 brain MRI cases from RadGenome-BrainMRI, where the latter consists of 55 diagnosis-with-grounding cases and 55 standard report-generation cases.

\vspace{3pt}
\noindent \textbf{Chest Imagenome for organ/anomaly grounding.}
Chest Imagenome~\cite{wu2021chest} is built on MIMIC-CXR~\cite{johnson2019mimic}, a de-identified collection of chest radiographs acquired in routine clinical practice at the Beth Israel Deaconess Medical Center (Boston, MA) between 2011 and 2016, so the images reflect real emergency-department acquisition conditions rather than a curated research protocol. It provides chest radiographs paired with structured scene-graph annotations derived from radiology reports and image-level/localized labels. Each scene graph links anatomical regions, such as lungs, cardiac silhouette, mediastinum, pleural regions, or other chest structures, with visual observations and abnormal findings. We use this structure to build organ/anomaly joint grounding cases. Specifically, we first keep cases with accessible chest X-ray images and non-empty local anatomical and abnormality annotations. We then map the original anatomical regions and observation labels to our benchmark schema, with anatomy set to chest, modality set to X-ray, and task type set to organ/anomaly joint grounding. The input to the agent preserves the original image and clinical question, while the reference answer is derived from the original localized anatomical and abnormality annotations. Cases are excluded when the target region is ambiguous, the abnormality label cannot be mapped to our task schema, or the required image/annotation pair is incomplete. After filtering, we randomly select 55 cases with diverse anatomical targets and abnormality types for the organ/anomaly grounding evaluation.

\vspace{3pt}
\noindent \textbf{RadGenome-BrainMRI for diagnosis and report generation.}
RadGenome-BrainMRI~\cite{Lei2025InterpretableBM} is used to construct the brain MRI part of the real-environment layer. It pairs multi-sequence brain MRI scans from public anomaly-segmentation cohorts with region-level abnormality masks and radiologist-written findings. We sample 110 cases in total from the BraTS-GLI, BraTS-MEN, and BraTS-MET cohorts, covering glioma, meningioma, and metastatic tumor cases. These cohorts are themselves multi-institutional: the BraTS collections aggregate pre-operative scans contributed by many sites and acquired under different clinical protocols and scanner models~\cite{baid2021rsna}, so the sampled cases carry acquisition variability across vendors, field strengths, and sequence parameters rather than originating from a single scanner or center. All selected cases are treated as head-and-neck MRI cases with 3D imaging inputs. We preserve the original 3D MRI data, tumor-related grounding annotations, and dataset-provided diagnostic or report references. Among the 110 cases, 55 are used for diagnosis with grounding clues, where the agent must combine imaging evidence and available grounding information to infer the disease category or clinically relevant diagnostic label. The remaining 55 cases are used for standard report generation, where the agent must synthesize a radiology-style response from the imaging input, tool outputs, and reference annotations. During sampling, we balance the cases across the three BraTS-derived cohorts as much as possible, while requiring complete image files, usable lesion annotations, and unambiguous task targets.

\vspace{3pt}
\noindent \textbf{Coverage and diversity of the real-environment layer.}
Taken together, the 165 cases span two imaging modalities (X-ray and MRI), two anatomical regions (chest and head \& neck), both 2D and 3D inputs, and two independent source collections drawn from different countries and care settings. Pathology coverage includes the localized chest abnormalities annotated in the Chest Imagenome scene graphs together with glioma, meningioma, and metastatic brain tumors. We nevertheless emphasize that this layer is designed for depth rather than breadth: it requires ground-truth annotations for both intermediate steps and final outcomes, and datasets with that level of annotation are scarce, so modality, anatomy, and pathology coverage is deliberately narrower than in the synthetic layer. Accordingly, the real-environment results should be read as a proof of concept on common modalities rather than as an estimate of agent performance across all of radiology; the two layers target different forms of representativeness, and their statistics are not directly comparable.

\begin{table*}[!t]
\centering
\caption{\textbf{Real-environment case sources.} The 165 real-world cases are collected from Chest Imagenome and RadGenome-BrainMRI without converting them into synthetic patient records.}
\label{tab:real_environment_case_sources}
\small
\setlength{\tabcolsep}{4pt}
\resizebox{\textwidth}{!}{%
\begin{tabular}{llllclp{0.32\linewidth}}
\toprule
\textbf{Source} & \textbf{Original Cohort} & \textbf{Task} & \textbf{Anatomy/Modality} & \textbf{Dim.} & \textbf{Cases} & \textbf{Reference Signal} \\
\midrule
Chest Imagenome & Chest X-ray scene graphs & Organ/anomaly joint grounding & Chest / X-ray & 2D & 55 & Local anatomical regions, abnormality labels, and grounding annotations from the original image--scene-graph pairs. \\
RadGenome-BrainMRI & BraTS-GLI, BraTS-MEN, BraTS-MET & Diagnosis with grounding clues & Head \& Neck / MRI & 3D & 55 & Tumor-related image annotations and diagnostic labels preserved from the original brain MRI cases. \\
RadGenome-BrainMRI & BraTS-GLI, BraTS-MEN, BraTS-MET & Standard report generation & Head \& Neck / MRI & 3D & 55 & Dataset-provided report/reference information and lesion annotations used as report-generation targets. \\
\midrule
\multicolumn{5}{r}{\textbf{Total}} & \textbf{165} & \\
\bottomrule
\end{tabular}%
}
\end{table*}

For automated tool-building evaluation, we further instantiate insufficient-tool scenarios from these real-case sources. Each scenario hides a required tool category, anatomy--modality capability, or functional requirement from the available tool set. The resulting task remains clinically meaningful because the original image and reference annotation are unchanged, but the initial tool environment becomes insufficient. This design allows us to test whether the agent can identify the missing capability, request an appropriate tool, and continue the workflow after tool construction.

\subsection{M3Builder-based Tool Pool Construction}
\label{sec:tool_pool}

We construct the executable tool pool following the M$^3$Builder protocol. The goal is not to use a single monolithic model, but to create a diverse set of specialized radiology tools with explicit tool cards and controllable performance differences. Each tool is associated with a task category, dataset source, anatomy, modality, dimensionality, input/output schema, supported targets, and performance metadata. These tool cards are exposed to the agent so that tool selection is based on declared capability and input/output requirements rather than hidden oracle information.

\vspace{3pt}
\noindent \textbf{Tool families and concrete tool variants.}
We build 14 dataset-specific tool families spanning four executable task groups: organ segmentation, anomaly detection, disease diagnosis, and report generation. For each tool family, we train or instantiate three versions using different data-subset configurations following the M$^3$Builder training protocol. These versions correspond to different training rounds or data scales and therefore have different empirical performance. Consequently, the pool contains $14\times3=42$ concrete executable tool variants. This design supports both the real-case evaluation and the simulated tool-set perturbations: the benchmark can provide only the strongest tool, provide multiple similar tools with different performance, remove a necessary capability, or include redundant distractor tools.

\begin{table*}[!t]
\centering
\caption{\textbf{M$^3$Builder tool families used for executable evaluation.} Each row denotes one dataset-specific tool family. For every family, three data-subset-trained variants are constructed and described with separate tool cards, yielding 42 concrete tool variants in total.}
\label{tab:m3builder_tool_pool}
\scriptsize
\setlength{\tabcolsep}{4pt}
\resizebox{\textwidth}{!}{%
\begin{tabular}{llllccp{0.30\linewidth}}
\toprule
\textbf{Tool Group} & \textbf{Dataset} & \textbf{Anatomy} & \textbf{Modality} & \textbf{Dim.} & \textbf{Variants} & \textbf{Primary Capability} \\
\midrule
Organ Segmentor & BTCV & Abdomen \& Pelvis & CT & 3D & 3 & Multi-organ abdominal and pelvic CT segmentation. \\
Organ Segmentor & Verse & Spine & CT & 3D & 3 & Vertebra/spine structure segmentation on CT. \\
Organ Segmentor & OASIS & Head \& Neck & MRI & 3D & 3 & Brain/head-and-neck structure segmentation on MRI. \\
\midrule
Anomaly Detector & COV19 & Chest & CT & 3D & 3 & Chest CT abnormality detection and localization. \\
Anomaly Detector & INS22 & Head \& Neck & CT & 3D & 3 & Head-and-neck CT anomaly localization. \\
Anomaly Detector & Panc & Abdomen \& Pelvis & CT & 3D & 3 & Pancreatic-region anomaly detection on CT. \\
Anomaly Detector & XDet10 & Chest & X-ray & 2D & 3 & Chest X-ray anomaly detection and localization. \\
\midrule
Imaging Diagnoser & ADNI & Head \& Neck & MRI & 3D & 3 & MRI-based neuroimaging disease diagnosis. \\
Imaging Diagnoser & KneeMR & Limb & MRI & 2D & 3 & Knee MRI disease or abnormality diagnosis. \\
Imaging Diagnoser & CC-CCII & Chest & CT & 3D & 3 & Chest CT disease diagnosis. \\
Imaging Diagnoser & KidCT & Abdomen \& Pelvis & CT & 2D & 3 & Kidney/abdominal CT disease diagnosis. \\
\midrule
Report Generator & CT-RATE & Chest & CT & 3D & 3 & Chest CT report generation. \\
Report Generator & GenomBra & Head \& Neck & MRI & 3D & 3 & Brain MRI report generation. \\
Report Generator & IU-Xray & Chest & X-ray & 2D & 3 & Chest X-ray report generation. \\
\midrule
\multicolumn{5}{r}{\textbf{Total concrete tool variants}} & \textbf{42} & $14$ tool families $\times$ $3$ subset-trained variants per family. \\
\bottomrule
\end{tabular}%
}
\end{table*}

\vspace{3pt}
\noindent \textbf{Three subset-trained versions per tool family.}
For each dataset-specific family, M$^3$Builder produces three variants trained or configured with different subsets of the available training data. We use these variants as low-, medium-, and high-performing tools under the same high-level capability. For example, three report-generation tools may all support chest CT report generation, but their performance scores differ because they correspond to different M$^3$Builder training rounds or data-subset sizes. This controlled performance diversity is important for the ``Differentiated'' setting, where the agent must choose the best-performing tool among several apparently suitable candidates. It is also useful for the ``Redundant'' setting, where multiple tools may share similar names or capabilities but differ in anatomy coverage, modality support, dimensionality, or input/output schema.

\vspace{3pt}
\noindent \textbf{Tool-card protocol.}
Every concrete tool variant is paired with a tool card following the M$^3$Builder protocol. Each card specifies: \textit{(1)} tool name and category; \textit{(2)} supported anatomy, modality, and image dimensionality; \textit{(3)} mandatory inputs, such as image, text query, or patient information; \textit{(4)} optional inputs, such as organ masks, anomaly masks, biomarkers, indicators, or previous tool outputs; \textit{(5)} output fields, such as masks, boxes, disease labels, reports, or treatment recommendations; \textit{(6)} capability constraints, including unsupported anatomies, modalities, or missing functions; and \textit{(7)} a performance score or rank estimated from the corresponding M$^3$Builder validation round. The agent only sees the tool cards and must infer whether a tool is usable from this structured description.

\vspace{3pt}
\noindent \textbf{Use in benchmark environments.}
The 42 M$^3$Builder variants are not all exposed simultaneously in every evaluation condition. Instead, different subsets are sampled to instantiate the benchmark environments described in the main text. The ``Baseline'' setting provides a minimal solvable set; the ``Redundant'' setting adds extra tools with overlapping or distractor capabilities; the ``Insufficient'' setting removes a required category, anatomy--modality coverage, or functional capability; and the ``Differentiated'' setting exposes multiple suitable tools with different performance scores. In the real-environment layer, the executable pool is restricted to tools relevant to the Chest Imagenome and RadGenome-BrainMRI tasks, with additional segmentation or diagnosis tools used when needed to match the clinical workflow. This separation ensures that the appendix-level tool construction details remain consistent with the main-text evaluation protocol.

\subsection{Layer-Specific AutoTB Protocols}
\label{sec:autotb_protocol}

\noindent \textbf{Synthetic-layer protocol.}
The verifier applies condition-specific checks that follow the three insufficient-tool configurations. Config1 requires the correct missing category; config2 requires anatomy and modality matching the case (or a valid universal scope); and config3 requires the affected category together with \texttt{InsufficientCapability}. Invalid or ungrounded requests fail the AutoTB criterion. A valid request produces a simulated tool card for the remainder of the workflow; it does not produce model weights, run training, or generate executable medical outputs. Thus, synthetic-layer results measure request grounding and idealized orchestration after verification.

\noindent \textbf{Real-environment workflow.}
The executable protocol invokes M$^3$Builder to construct a candidate tool, validates the candidate against the requested tool card, registers a successful candidate in the tool pool, and then replans and executes the augmented workflow. The logs preserve construction status, validation outcome, and downstream invocation outcome, allowing construction failures to be separated from incorrect tool usage. Ungrounded or unverifiable requests are returned as grounded refusals. Algorithm~\ref{alg:autotb} describes this real-environment protocol.

\begin{lstlisting}[
    backgroundcolor=\color{lightgray!20},
    basicstyle=\footnotesize\ttfamily,
    breaklines=true,
    frame=single,
    columns=flexible,
    caption={Pseudo-code of executable AutoTB in the real-environment layer.},
    label={alg:autotb}
]
Input: task T, tool cards U, context protocol P, max trials K
Output: final answer y or grounded refusal r

1  z <- Plan(T | U)
2  if Executable(z, U) then return Run(z)
3  m <- GroundMissingCapability(T, z, U)
4  if not Grounded(m) then return Refuse(m)
5  g <- SpecifyToolCard(m)   # category, anatomy, modality, I/O
6  C0, R0 <- empty
7  for k = 1 ... K do
8      Ck, Rk <- M3Builder(Ck-1, Rk-1, g | P)
9      if Validate(Ck, g) then
10         U <- U union {Tool(Ck, g)}
11         return Run(Replan(T | U))
12 return Refuse(RK)
\end{lstlisting}

\noindent \textbf{Representative implementation traces.}
The following traces instantiate the control flow and field schemas in the released RadA-BenchPlat and M$^3$Builder implementations together with the evaluation protocol above. They make the decision and attribution rules auditable and are not additional observations used to calculate the aggregate results in the main paper.

\begin{lstlisting}[
    backgroundcolor=\color{lightgray!20},
    basicstyle=\scriptsize\ttfamily,
    breaklines=true,
    frame=single,
    columns=flexible,
    caption={Synthetic-layer AutoTB request-verification trace (config2).},
    label={alg:autotb_synthetic_trace}
]
[CASE]
layer=synthetic
condition=Insufficient-config2
anatomy=Chest; modality=X-ray
required_step=Generate a radiology report
available_matching_tool=None

[AGENT OUTPUT]
<NoCall>
 <Purpose>Generate a report for the chest X-ray</Purpose>
 <Category>Report Generator</Category>
 <Anatomy>Chest</Anatomy>
 <Modality>X-ray</Modality>
 <Ability>SpecificToolMissing</Ability>
</NoCall>

[PARSER]
flag=NoCall
fields={Category: Report Generator, Anatomy: Chest,
        Modality: X-ray, Ability: SpecificToolMissing}

[VERIFIER]
has_deny=True
matches_anatomy=True
matches_modality=True
request_status=ACCEPTED
autotb_success=1

[POOL UPDATE]
category=Report Generator
property=Chest X-ray report generator
compulsory_input=[$Image$]
output=[$Report$]
training_or_model_weights=NOT_RUN
next_action=Replan with the verified tool card
\end{lstlisting}

\begin{lstlisting}[
    backgroundcolor=\color{lightgray!20},
    basicstyle=\scriptsize\ttfamily,
    breaklines=true,
    frame=single,
    columns=flexible,
    caption={Representative real-environment AutoTB protocol trace.},
    label={alg:autotb_real_trace}
]
[AUTOTB REQUEST]
task=Report Generation
anatomy=Chest; modality=X-ray; dimensionality=2D
required_input=Image; required_output=Report
human_requirements="Train a model for report generation using 2D chest X-ray images."

[M3BUILDER: DATASET SELECTOR]
selected_dataset=IU-Xray
selection_basis=2D X-ray Chest report generation

[M3BUILDER: DATA PROCESSOR]
reference_schema=report_generation
outputs=[train.json, test.json]
status=COMPLETED

[M3BUILDER: DATALOADER]
template=ReportGeneration_Template.py
output=TrainPipeline/Datapath/dataloader.py
full_iteration_check=PASSED

[M3BUILDER: MODEL TRAINER]
template=ReferenceFiles/TrainingScripts/ReportGeneration
command=torchrun ... train_report.py ... 2>&1 | tee output.log
stdout_sequence=[Setup Data, Setup Model, Setup Trainer,
                 Start Training, trainer.evaluate]
required_artifacts=[output.log, checkpoint, evaluation metrics]
validation=PASSED

[TOOL REGISTRATION]
category=Report Generator
property=Chest X-ray report generator
compulsory_input=[$Image$]
output=[$Report$]
build_status=SUCCESS; registry_status=ACTIVE

[DOWNSTREAM INVOCATION]
call=<EndCall><Tool>AutoTB-RG</Tool><Input>[$Image$]</Input></EndCall>
input_schema_check=PASSED
runtime_status=SUCCESS
attribution=SUCCESSFUL_BUILD_AND_USE
\end{lstlisting}

\noindent \textbf{Failure attribution from the trace.}
A failure before a validated candidate exists---for example, dataset selection failure, a non-executable dataloader, training termination, a missing checkpoint, or absent evaluation output---is assigned to the construction side and the candidate is not registered. Environment and dependency failures are recorded separately from a candidate that completes training but lacks the requested functionality. By contrast, if validation and registration succeed but the downstream call supplies an incompatible input, omits a compulsory input, or requests an unsupported anatomy/modality, the failure is attributed to \textit{Usage Error}. If the agent never emits the required \texttt{NoCall}/build request, it is attributed to \textit{Planning Error}. These event boundaries operationalize the categories reported in Table~\ref{tab:autotb_failure_modes}.

\subsection{Feature Distributions in Patient Record}

\begin{figure*}[h]
    \centering
    \includegraphics[width=1\linewidth]{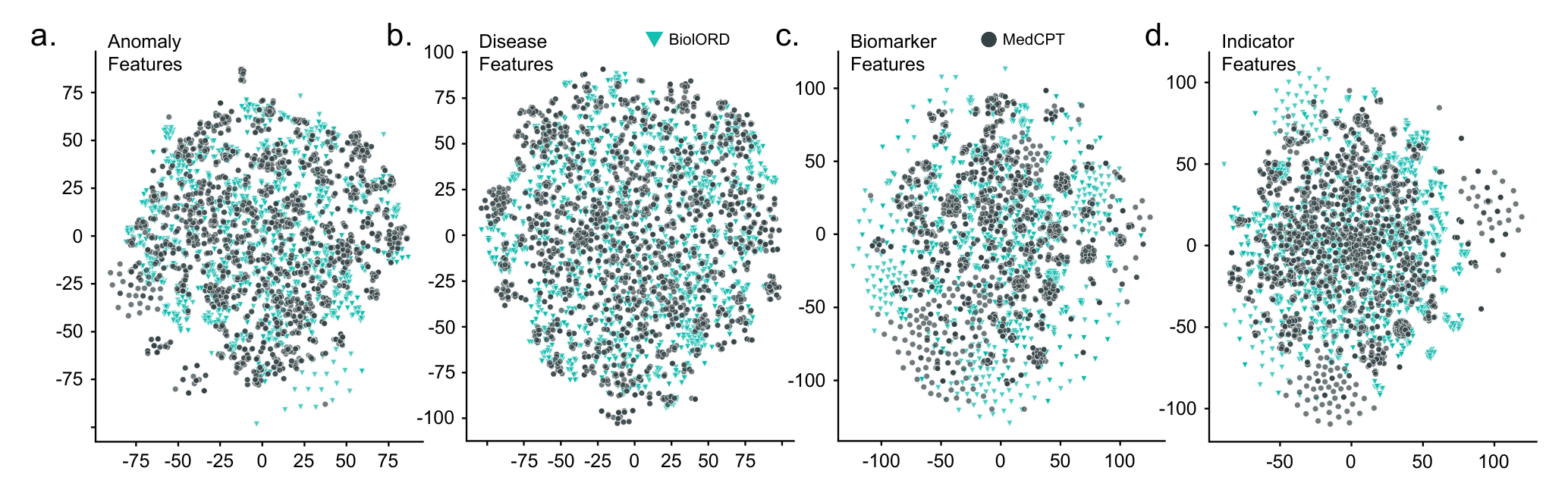}
    \vspace{1mm}
    \caption{Distributions of anomaly, disease, biomarker, and indicator features extracted using BioLORD and MedCPT.}
    \label{fig:Distribution}
\end{figure*}

Extended Figure~\ref{fig:Distribution}e visualizes anomalies, diseases, biomarkers, and indicators within patient records. Here, ``biomarkers'' refer to imaging features ({\em e.g.}, dimensions and textures), while ``indicators'' correspond to clinical classifications ({\em e.g.}, cancer staging) or scoring systems ({\em e.g.}, CURB-65). We embed these attributes using BioLORD and MedCPT. The t-SNE plot shows broad coverage across clinical scenarios.

\subsection{Context Token Consumption Analysis}

\begin{figure}[h]
    \centering
    \includegraphics[width=1\linewidth]{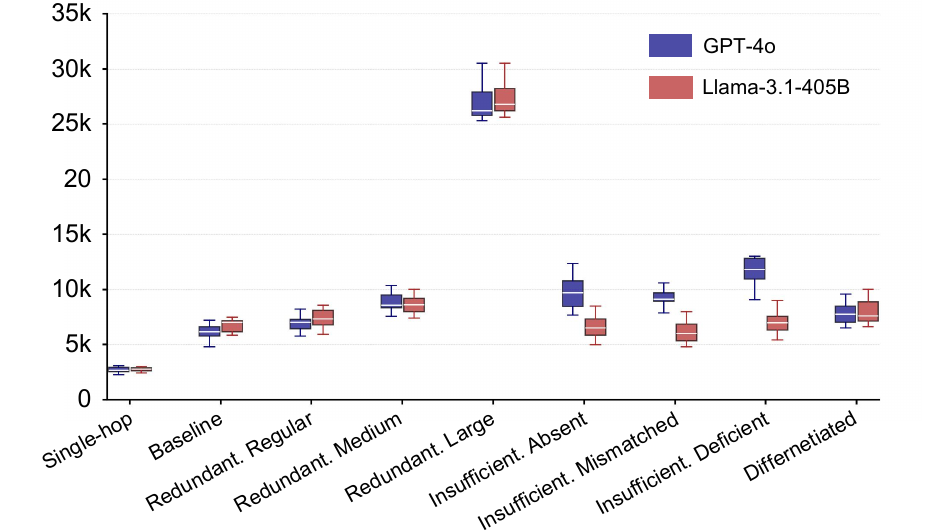}
    \vspace{1mm}
    \caption{Context token length in 8 simulated conditions using GPT-4o/Llama-3.1-405B as the agent core.}
    \label{fig:Token}
\end{figure}

As shown in Extended Figure~\ref{fig:Token}, we present the multi-turn response token lengths generated by two LLMs, GPT-4o~\cite{openai2024gpt4o} and Llama-3.1~\cite{grattafiori2024llama3}, representing closed-source and open-source models, respectively. In most conditions, the total token lengths for both models are comparable, ranging from 3,000 to 30,000 tokens.
However, in the Insufficient tool set conditions, GPT-4o exhibits significantly longer context lengths than Llama, primarily due to Llama's frequent failure to execute proper denial responses, resulting in premature terminations. 

\begin{figure*}[!h]
    \centering
    \includegraphics[width=1\linewidth]{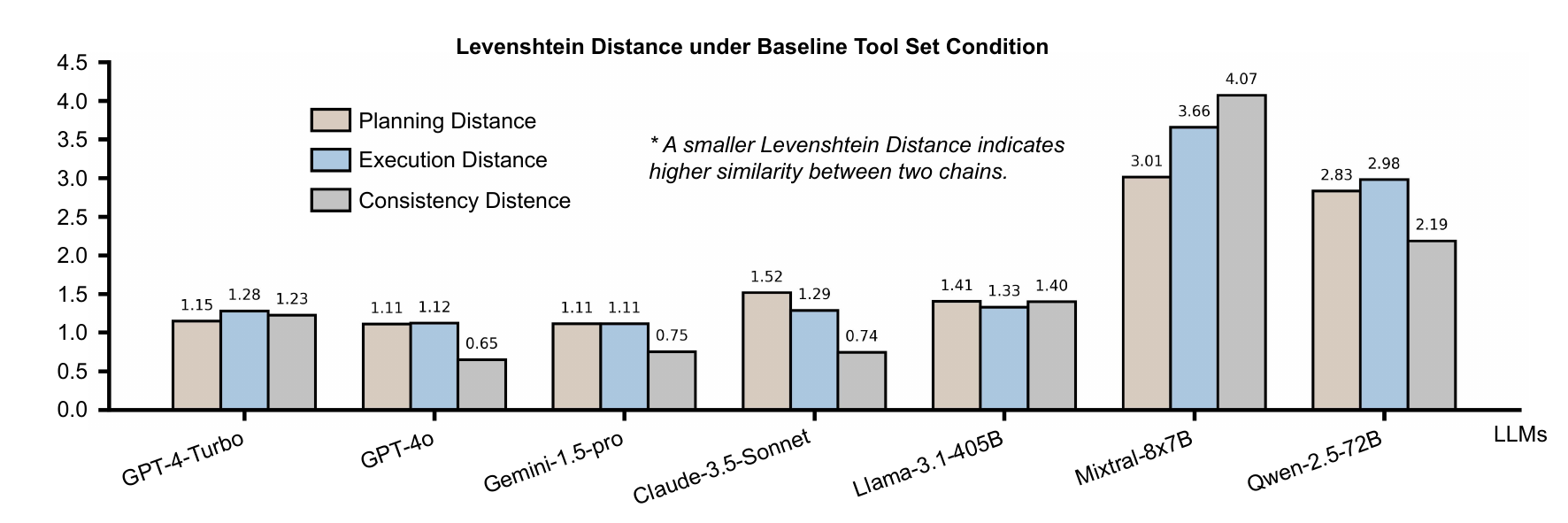}
    \vspace{1mm}
    \caption{Three Levenshtein distances between the planned, executed, and ground-truth tool chains under the baseline condition for seven LLMs.}
    \label{fig:Baseline}
\end{figure*}

\subsection{Performance on Baseline Tool Set Condition}

\textbf{1)} Closed-source models generally outperform open-source models (except Llama-3.1-405B) under the baseline tool-set condition. As shown in Extended Figure~\ref{fig:Baseline}, their planning-to-ground-truth LD is below 1.5, indicating fewer than 1.5 edits on average for chains whose mean length is five steps.
\textbf{2)} Differences emerge between planned and executed chains during iterative execution, as models may revise their tool selections. Claude-3.5 and Llama-3.1 move closer to the ground-truth chains, whereas the other models maintain or increase their deviations.

\subsection{A Patient Record Example}
Here, we present a concrete patient record from the synthetic layer of RadA-BenchPlat:

\begin{lstlisting}[
    backgroundcolor=\color{lightgray!20},
    basicstyle=\small\ttfamily,
    breaklines=true,
    frame=single,
    breakatwhitespace=true,
    columns=flexible,
    label=lst:patient_record_example,
    caption=Patient Record Example
]
"Information": {
    "Age": "42",
    "Sex": "Female",
    "Height": "165",
    "Weight": "68",
    "History": "Patient has a history of seasonal allergies and recurrent upper 
    respiratory infections",
    "Complaint": "Persistent facial pain, nasal congestion, and headache for the 
    past 2 weeks"
},
"Anatomy": "Head and Neck",
"Modality": "X-ray",
"Anomaly": {
    "Part": "Maxillary sinuses",
    "Symptom": "Opacification"
},
"Disease": "Sinusitis",
"OrganBiomarker": {
    "OrganObject": "Maxillary sinus",
    "OrganDim": "density",
    "OrganQuant": "+40 Hounsfield Units"
},
"AnomalyBiomarker": {
    "AnomalyObject": "Opacification",
    "AnomalyDim": "intensity",
    "AnomalyQuant": "80% increase compared to normal airspace"
},
"Indicator": {
    "Name": "Lund-Mackay Score",
    "Value": "8 (Moderate sinusitis)"
},
"Report": {
    "Finding": "X-ray of the paranasal sinuses demonstrates bilateral maxillary 
    sinus opacification. The right maxillary sinus shows complete opacification, 
    while the left maxillary sinus demonstrates air-fluid levels. Frontal and ethmoid 
    sinuses appear clear. No evidence of bone erosion or destruction. Nasal septum 
    appears midline. Soft tissues of the face and neck are unremarkable.",
    "Impression": "Findings consistent with bilateral maxillary sinusitis, more 
    pronounced on the right side. No evidence of complications such as orbital or 
    intracranial involvement."
},
"Treatment": "Given the patient's symptoms and radiographic findings, a diagnosis 
of acute bacterial sinusitis is likely. Initial treatment should include a 10-14 day 
course of broad-spectrum antibiotics such as amoxicillin-clavulanate or, in case of 
penicillin allergy, a respiratory fluoroquinolone. Adjunctive treatments include nasal 
saline irrigation, intranasal corticosteroids, and oral decongestants for symptom 
relief. The patient should be advised to stay well-hydrated and use over-the-counter 
pain relievers as needed. If symptoms persist or worsen after 72 hours of antibiotic 
therapy, reassessment is warranted. A follow-up appointment should be scheduled in 2-3 
weeks to ensure resolution of symptoms. If recurrent episodes occur, further evaluation 
with CT imaging and potential referral to an ENT specialist for consideration of 
endoscopic sinus surgery may be necessary."

\end{lstlisting}

\vspace{3pt}
which is generated by prompting GPT-4 leveraging the following prompts:
\begin{lstlisting}[
    backgroundcolor=\color{lightgray!20},
    basicstyle=\small\ttfamily,
    breaklines=true,
    frame=single,
    breakatwhitespace=true,
    columns=flexible,
    label=lst:patient_record_generation_prompt,
    caption=Patient Record Generation Prompt
]

You are an experienced clinical radiologist. Your task is to generate a detailed 
medical case based on a hypothetical 256x256 medical image. I will provide you with 
the Anatomy and Modality of the image, as well as an overall Disease. Using this 
information, you should create a comprehensive case report including patient 
information, specific anomalies, biomarkers, indicators, a radiology report, and 
treatment recommendations.

Please structure your response using the following template:

<Case>
    <Information>
        <Age> [Number without units] </Age>
        <Sex> [Male / Female] </Sex>
        <Height> [Number in cm] </Height>
        <Weight> [Number in kg] </Weight>
        <History> [Brief descriptive text] </History>
        <Complaint> [Brief descriptive text] </Complaint>
    </Information>
    <Anatomy> [Head and Neck / Chest / Breast / Abdomen and Pelvis / Limb / Spine] 
    </Anatomy>
    <Modality> [CT / MRI / X-ray / Ultrasound / Mammography] </Modality>
    <Anomaly>
        <Part> [Specific location of anomaly (e.g., right upper lobe of lung)] </Part>
        <Symptom> [Type of anomaly (e.g., nodule)] </Symptom>
    </Anomaly>
    <Disease> [Corresponding disease name] </Disease>
    <OrganBiomarker>
        <OrganObject> [A specific organ serving as biomarker] </OrganObject>
        <OrganDim> [number / length / size / volume / angle / density / intensity / 
        texture] </OrganDim>
        <OrganQuant> [Specific quantitative value] </OrganQuant>
    </OrganBiomarker>
    <AnomalyBiomarker>
        <AnomalyObject> [The same Anomaly described in the Anomaly Symptom serving 
        as biomarker] </AnomalyObject>
        <AnomalyDim> [number / length / size / volume / angle / density / intensity / 
        texture] </AnomalyDim>
        <AnomalyQuant> [Specific quantitative value] </AnomalyQuant>
    </AnomalyBiomarker>
    <Indicator>
        <Name> [Name of the indicator (e.g., Lung Cancer TNM Staging Score)] </Name>
        <Value> [Specific value or grade (e.g., cT2aN0M0 (Stage IB))] </Value>
    </Indicator>
    <Report>
        <Finding> [Findings section in the style of a MIMIC-CXR report] </Finding>
        <Impression> [Impression section in the style of a MIMIC-CXR report] 
        </Impression>
    </Report>
    <Treatment> [A paragraph including diagnostic procedures, medication 
    recommendations, and follow-up suggestions] </Treatment>
</Case>

Guidelines for generating the case report:

Information: Provide realistic patient demographics, medical history, and chief 
complaint.
Anomaly: Describe a specific anomaly consistent with the given anatomy, modality, 
and disease.
Biomarkers: Choose a relevant organ biomarker and the anomaly biomarker mentioned 
before that can be observed in the image and provide plausible quantitative value.
Indicator: Calculate an appropriate indicator based on the patient information and 
biomarker value. Provide a specific, medically accurate score or grade.
Report: Generate a concise radiological report in the style of MIMIC-CXR, with 
separate Findings and Impression sections.
Treatment: Offer a comprehensive treatment plan including diagnostic procedures, 
medication recommendations, and follow-up care.
Ensure that all parts of the case are medically accurate and consistent with each 
other. Use your expertise as a radiologist to provide realistic and detailed 
information throughout the case.

Given the following parameters:
- Anatomy: {ANATOMY}
- Imaging Modality: {MODALITY}
- Disease: {DISEASE}
Please generate this patient record.
\end{lstlisting}

\subsection{A Tool Card Example}

Here is an example of a tool card of a report generation tool focused on Head and Neck X-ray images:

\begin{lstlisting}[
    backgroundcolor=\color{lightgray!20},
    basicstyle=\small\ttfamily,
    breaklines=true,
    frame=single,
    breakatwhitespace=true,
    columns=flexible,
    label=lst:tool_card_example,
    caption=Tool Card Example
]
=== Tool Description for TOOL17 ===
Name: TOOL17
Category: Report Generator
Ability: Given the Head and Neck X-ray Image, any other text information and 
organ/anomaly masks and labels, generate a radiology report.
Property: Report Generator only suitable for Head and Neck X-ray image with 
Text and Mask
Compulsory Input: [`$Image$']
Optional Input: [`$Information$', `$OrganObject$', `$AnomalyObject$', `$Disease$', 
`$OrganDim$', `$OrganQuant$', `$AnomalyDim$', `$AnomalyQuant$', `$IndicatorName$', 
`$ValueName$', `$OrganMask$', `$AnomalyMask$']
Output: [`$Report$']
Performance: Score from 0.4 to 0.88, increases with optional inputs
\end{lstlisting}

\subsection{A Task Simulation Example}
Here is an example of QA pair generation from a chest X-ray record of a pneumonia patient:

\begin{lstlisting}[
    backgroundcolor=\color{lightgray!20},
    basicstyle=\small\ttfamily,
    breaklines=true,
    frame=single,
    breakatwhitespace=true,
    columns=flexible,
    label=lst:qa_pair_example,
    caption=QA Pair Example
]
    <Q1> Identify and segment the lung fields in this chest X-ray. </Q1>
    <A1> The lung fields are segmented [{\em organ mask}: left lung, right lung]. 
    Both lungs show clear boundaries. </A1>

    ...

    <Q11> Based on the imaging findings and clinical indicators, what treatment plan 
    would you recommend? </Q11>
    <A11> Given the moderate pneumonia severity (CURB-65 score: 2) and [anomaly mask], 
    recommend oral antibiotics and follow-up chest X-ray in 2 weeks. </A11>
    
\end{lstlisting}

which is generated by the following prompt, leveraging GPT-4:

\begin{lstlisting}[
    backgroundcolor=\color{lightgray!20},
    basicstyle=\small\ttfamily,
    breaklines=true,
    frame=single,
    breakatwhitespace=true,
    columns=flexible,
    label=lst:qa_pair_generation_prompt,
    caption=QA Pair Generation Prompt
]
Assume a clinical medical imaging scenario where you, as the Agent core, play the role 
of a doctor. Given a patient's radiological image, you want to complete different tasks 
by calling various tools. There are ten tools in total (numbered 0 to 9):

TOOLKIT 
0. |Modality Classifier|
    Property: A classification model
    Ability: Determine the modality of the Image.
    Input: [Image]
    Output: [Modality]

1. |Anatomy Classifier|
    Property: A classification model,
    Ability: Determine the anatomy of the Image.
    Input: [Image]
    Output: [Anatomy]

2. |Organ Segmentor|
    Property: A segmentation model
    Ability: Given the modality and anatomy, segment all organs in the Image 
    (can not segment any lesion or abnormality).
    Input: [Image] & [Modality] & [Anatomy]
    Output: [Organ Mask] & [Organ Label]

3. |Anomaly Detector|
    Property: A detection model
    Ability: Given the modality and anatomy, determine the location and type of 
    abnormality.
    Input: [Image] & [Modality] & [Anatomy]
    Output: [Anomaly Mask] & [Anomaly Label]

4. |Imaging Diagnoser|
    Property: A classification model
    Ability: Given the modality and anatomy, diagnose diseases directly from the input 
    image.
    Input: [Image] & [Modality] & [Anatomy]
    Output: [Diseases]

5. |Grounded Diagnoser|
    Property: An inference model
    Ability: Diagnose diseases based on organ segmentation and anomaly detection results.
    Input: [Image] & [Organ Mask] & [Organ Label] & [Anomaly Mask] & [Anomaly Label]
    Output: [Diseases]

6. |Biomarker Quantifier|
    Property: A quantification model
    Ability: Given the organ region or anomaly region and the biomarker of interest, 
    estimate its property. The biomarker can be either organ or anomaly. The dimension 
    can be one of the number, length, size, volume, angle, density, intensity or 
    texture of the organ or anomaly.
    Input: [Image] & [Object] & [Dim] & [Biomarker Mask] & [Biomarker Label]
    Output: [Quant]

7. |Indicator Calculator|
    Property: A calculation model
    Ability: Use prior patient information and several biomarkers values to calculate 
    the indicator, the indicator can be a score or a grading.
    Input: [Priors] & [Indicator] & [Biomarkers] & [Quants]
    Output: [Value]

8. |Report Generator|
    Property: A multimodal model
    Ability: Generate a medical report by integrating results processed by former tools.
    Input: [Image] & [Modality] & [Anatomy]
    Optional Input: ([Organ Mask] & [Organ Label]), ([Anomaly Mask] & [Anomaly Label]), 
    [Diseases], ([Object] & [Dim] & [Quant]), ([Indicators] & [Values])
    Output: [Report]

9. |Treatment Planner|
    Property: a language model
    Ability: Recommend personalized treatment plans based on all results processed by 
    former tools and the patient's information.
    Input: [Image] & [Priors] & [Modality] & [Anatomy] & [Diseases]
    Optional Input: ([Organ Mask] & [Organ Label]),([Anomaly Mask] & [Anomaly Label]), 
    ([Object] & [Dim] & [Quant]), ([Indicators] & [Values])
    Output: [Treatment]

There are 11 different tasks and the chain of tools each task hopes to break down into:

(1) Basic Image Analysis and Organ Segmentation
    ToolUse: 012
    Description: Perform basic image analysis and segment organs within the medical 
    image.

(2) Basic Image Analysis and Anomaly Detection
    ToolUse: 013
    Description: Perform basic image analysis and detect anomalies within the medical 
    image.

(3) Image-based Direct Disease Diagnosis
    ToolUse: 014
    Description: Diagnose disease directly from the medical image without intermediate 
    steps.

(4) Organ segmentation and anomaly localization
    ToolUse: 0123
    Description: Segment organs and locate anomalies within the medical image.

(5) Anomaly-based Disease Diagnosis
    ToolUse: 01235
    Description: Diagnose disease based on grounded diagnosis by finding abnormalities 
    and the organ in which they occur.

(6) Organ Biomarker Quantification  
    ToolUse: 0126
    Description: Quantify specific biomarkers related to organs in the medical image.

(7) Anomaly Biomarker Quantification
    ToolUse: 0136
    Description: Quantify specific biomarkers related to anomalies in the medical image.

(8) Disease and Anomaly Based Report Generation
    ToolUse: 01348
    Description: Generate a medical report based on detected diseases and anomalies.

(9) Disease and Biomarker Based Report Generation
    ToolUse: 0123568
    Description: Generate a comprehensive report incorporating anomaly detection, 
    disease diagnosis and biomarker quantification.

(10) Comprehensive Evaluation Report Generation
    ToolUse: 01235678
    Description: Generate a detailed evaluation report including all aspects of the 
    medical image analysis.

(11) Comprehensive Report Generation and Treatment Planning
    ToolUse: 012356789
    Description: Generate a comprehensive report including all analysis results and 
    treatment plans.

Please generate 11 mutually independent question-answer pairs corresponding to tasks 
above, based on the different task natures and the content of the case. Specifically, 
strictly avoid including information 
in the questions that should be determined by the tools (such as imaging modality, 
specific anatomy, or precise abnormality types).
Pay attention: create task-specific question-answer pairs that highlight the unique 
tool usage patterns for different tasks. The questions should be:

1. Naturally aligned with the task description
2. Representative of real-world scenarios for that task type
3. Questions must not reveal information about modality or anatomy that should be 
determined by Tools 0 and 1 or other tools. The questions should be phrased in a way 
that necessitates the use of these basic identification tools.

Provide answers in the most concise free-text form possible. If visual results such as 
masks are involved, please embed them in the text in the form of [Organ Mask] or 
[Anomaly Mask]. (eg. The organ segmentation result is shown as [Organ Mask].) The 11 
generated question-answer pairs should follow this template:

<Q1> ... </Q1>
<A1> ... </A1>
<Q2> ... </Q2>
<A2> ... </A2>
...
<Q11> ... </Q11>
<A11> ... </A11>

The answer only needs to provide a simple final result based on the information already 
available in the Case, without showing the thought process. I will now provide you with 
a Case containing all the information.

The patient record is {Patient Record}, please generate the corresponding QA pairs.
    
    
\end{lstlisting}

\subsection{Detailed Workflow Prompts in RadA-BenchPlat}
In this part we detail the prompt template used in the main workflow of RadA-BenchPlat.

\vspace{3pt}
\noindent\textbf{Task Decomposition.}

\begin{lstlisting}[
    backgroundcolor=\color{lightgray!20},
    basicstyle=\small\ttfamily,
    breaklines=true,
    frame=single,
    breakatwhitespace=true,
    columns=flexible,
    label=lst:task_decomposition_prompts,
    caption=Task Decomposition Prompts
]
# Medical Image Analysis Assistant

## Task Overview
You are a radiological agent analyzing medical images. 
For each query, you will receive:
    1. A medical imaging examination (Image) of a patient (assume already provided)
    2. Known patient Information including demographics, medical history, 
    and main complaints.

Your task involves three sequential parts:

1. Problem Decomposition (Part 1)
- Identify available information
- Break down the question into sequential steps

2. Sequential Tool Application (Part 2)  
- Execute one tool at a time
- Record each tool's output
- Continue until sufficient information is gathered

3. Solution Synthesis (Part 3)
- Integrate all results
- Generate final answer

## Available Information Categories
The following categories must be used exactly as written:

['$Information$', '$Anatomy$', '$Modality$', '$Disease$', '$OrganObject$', 
'$OrganDim$', '$OrganQuant$', '$AnomalyObject$', '$AnomalyDim$', 
'$AnomalyQuant$', '$IndicatorName$', '$IndicatorValue$', '$Report$', '$Treatment$']

Where:
- $Information$: Patient demographics 
(e.g., "45-year-old male", "BMI: 24", "history of diabetes")
- $Anatomy$: Body part (e.g., "chest", "abdomen", "brain")
- $Modality$: Imaging technique (e.g., "X-ray", "CT", "MRI")
- $Disease$: Medical condition (e.g., "pneumonia", "cancer", "fracture") 
- $OrganObject$: Organ to measure (e.g., "liver", "heart")
- $OrganDim$: Organ measurement type (e.g., "number", "length", "size", 
  "volume", "angle", "density", "intensity", "texture")
- $OrganQuant$: Organ measurement value (e.g., "5cm", "120ml")
- $AnomalyObject$: Abnormality to measure (e.g., "tumor", "fracture")
- $AnomalyDim$: Abnormality measurement type (e.g., "number", "length", "size", 
  "volume", "angle", "density", "intensity", "texture")
- $AnomalyQuant$: Abnormality measurement value (e.g., "2cm", "5ml")
- $IndicatorName$: Clinical indicator name
- $IndicatorValue$: Clinical indicator value
- $Report$: Medical report content
- $Treatment$: Treatment recommendations

## Available Tool Categories
Tool categories must be used exactly as written:

[*Anatomy Classifier*, *Modality Classifier*, 
*Organ Segmentor*, *Anomaly Detector*, 
*Imaging Diagnoser*, *Grounded Diagnoser*, 
*Biomarker Quantifier*, *Indicator Calculator*, 
*Report Generator*, *Treatment Planner*]

## Response Format for Part 1
For each query, respond ONLY with:

Known Info: [list any categories explicitly mentioned in the query]
Tool Chain: [list required tools connected by ->]

## Examples

Query 1: "For a straightforward approach to diagnosing the patient's condition 
based on her symptoms and the image, what diseases can be directly identified?"
Response:
Known Info: []
Tool Chain: [*Anatomy Classifier* -> *Modality Classifier* 
             -> *Imaging Diagnoser*]

Query 2: "This 45-year-old male's chest CT shows a 2cm nodule in the right lung. 
Can you give a report?"
Response:
Known Info: ['$Information$', '$Anatomy$', '$Modality$', '$AnomalyObject$', 
            '$AnomalyDim$', '$AnomalyQuant$']
Tool Chain: [*Organ Segmentor* -> *Anomaly Detector* 
             -> *Grounded Diagnoser* -> *Report Generator*] 
             (because some information is provided, so 
             *Anatomy Classifier*, *Modality Classifier*, 
             *Biomarker Quantifier* are optimized.)

## Important Rules
1. Assume the medical image is already provided
2. Use exact item category names with $$ as listed (e.g., '$Anatomy$')
3. Use exact tool category names with ** as listed 
   (e.g., '*Anatomy Classifier*')
4. Only respond with Part 1 analysis - Parts 2 & 3 will be addressed 
   in subsequent interactions
5. Include only the categories explicitly mentioned in the query
6. Connect tools using -> symbol

Please wait for my query. 
When provided, analyze it following the format shown in the examples above.

{Patient Record}

{Query}
    
\end{lstlisting}

\vspace{3pt}
\noindent\textbf{Tool Selection \& Execution. }
\begin{lstlisting}[
    backgroundcolor=\color{lightgray!20},
    basicstyle=\small\ttfamily,
    breaklines=true,
    frame=single,
    breakatwhitespace=true,
    columns=flexible,
    label=lst:tool_selection_execution_prompts,
    caption=Tool Selection \& Execution Prompts
]
# Next Step Planning

## Current Status
Current results dictionary: {value_dict}

## Planning Guidelines
1. Reference your high-level tool chain from Part 1 decomposition
2. Consider current results to refine specific tool selection
3. Maintain sequential progression according to planned workflow
4. Adjust tool selection if needed based on intermediate results
5. Check if the tool category is missing when this category of tools is required
6. Check if the tool is suitable for the detected Anatomy and Modality in reserved 
value dictionary based on the Tool description Ability and Property
7. Check if the result in reserved value dictionary can be derived from each tool used 
in former steps based on the limited label list described in Tool Ability 
8. If no suitable tool exists, identify which type of denial applies:
   - Missing tool category
   - Missing specific modality-anatomy tool
   - Insufficient tool capability

## Input Requirements
1. Required inputs: Must include all mandatory inputs specified in tool description
2. Optional inputs: Include if available and beneficial to tool performance
3. Do not include variables that are not relevant to the tool's function
4. All variables must exist in the current results dictionary
5. Use proper $$ notation for all variables

## Response Format
For ongoing analysis (if not final step):
<Call>
    <Purpose>Brief, clear statement of this step's goal in context of overall 
    analysis</Purpose>
    <Tool>TOOL[number] - must match available specific tools</Tool>
    <Input>['$variable1$', '$variable2$', ...] - use only existing variables from 
    results</Input>
</Call>

For final step only:
<EndCall>
    <Purpose>Brief, clear statement of this final step's goal</Purpose>
    <Tool>TOOL[number] - must match available specific tools</Tool>
    <Input>['$variable1$', '$variable2$', ...] - use only existing variables from 
    results</Input>
</EndCall>

When a tool category is completely missing:
<NoCall>
    <Purpose>The purpose requiring a missing tool category</Purpose>
    <Category>The missing category from ['Anatomy Classifier', 'Modality Classifier', 
    'Organ Segmentor', 'Anomaly Detector', 'Imaging Diagnoser', 'Grounded Diagnoser', 
    'Biomarker Quantifier', 'Indicator Calculator', 'Report Generator', 'Treatment 
    Planner']</Category>
    <Anatomy>Universal</Anatomy>
    <Modality>Universal</Modality>
    <Ability>CategoryMissing</Ability>
</NoCall>

When specific tools for a modality-anatomy combination are missing:
<NoCall>
    <Purpose>The purpose requiring a specific modality-anatomy tool</Purpose>
    <Category>The required tool category</Category>
    <Anatomy>The specific anatomy from ['Universal', 'Head and Neck', 'Chest', 
    'Breast', 'Abdomen and Pelvis', 'Limb', 'Spine']</Anatomy>
    <Modality>The specific modality from ['Universal', 'X-ray', 'CT', 'MRI', 
    'Ultrasound']</Modality>
    <Ability>SpecificToolMissing</Ability>
</NoCall>

When existing tools lack required capabilities:
<NoCall>
    <Purpose>The purpose requiring advanced capabilities</Purpose>
    <Category>The category of existing but insufficient tools</Category>
    <Anatomy>The relevant anatomy</Anatomy>
    <Modality>The relevant modality</Modality>
    <Ability>InsufficientCapability</Ability>
</NoCall>

## Format Requirements
1. Maintain proper XML structure
2. Use exact tool numbers as specified in tool descriptions
3. Mark all variables with $$ notation
4. Include only existing variables from results dictionary
5. Keep purpose statements clear and concise
6. Brief response only includes one Call, EndCall, or NoCall XML part without 
additional explanations
7. For NoCall responses, use the appropriate format based on denial type

## Decision Making Process
1. Review planned tool chain from Part 1
2. Check current results in value dictionary
3. Check if tool category is missing when this category of tools is required
4. Check tool Ability and Property in detail to judge its suitability for detected 
Anatomy and Modality in the value dictionary
5. Check if the result in reserved value dictionary can be derived from each tool used 
in former steps based on the limited label list described in Tool Ability
6. Evaluate tool availability and capability:
   - Is the required tool category available?
   - Are specific tools available for the needed modality-anatomy combination?
   - Do available tools have sufficient capabilities?
7. If tools are insufficient, use appropriate NoCall format
8. If tools are available, select and format appropriate Call/EndCall
9. Use <EndCall> only if this will be the final step

Please provide your next step based on:
- Original tool chain plan
- Current results
- Available specific tools
- Remaining analysis needs
- Tool availability and capability assessment
    
\end{lstlisting}
\vspace{3pt}
\noindent\textbf{Response Generation.}
\begin{lstlisting}[
    backgroundcolor=\color{lightgray!20},
    basicstyle=\small\ttfamily,
    breaklines=true,
    frame=single,
    breakatwhitespace=true,
    columns=flexible,
    label=lst:response_generation_prompt,
    caption=Response Generation Prompt
]

Based on your Part 1 analysis plan, Part 2 tool execution sequence, and the final 
results dictionary {value_dict}, provide:

    1. A concise answer to the initial question
    2. Key supporting evidence from your results
    3. How your findings align with the planned analysis

Keep your response brief and focused on directly answer the initial question.
    
\end{lstlisting}

\subsection{An In-depth Case Study}

Here we select a representative case from the agent core evaluation workflow to demonstrate the process, where we test the agent cores' performance using a QA pair based on the task type "Anomaly-wise Biomarker Calculation" and a patient record (a 62-year-old female with hypertension, osteoarthritis of the knees, and chronic neck pain with stiffness radiating to right shoulder) under the NR.~Denyl2 tool set condition, using Claude-3.5-Sonnet as the agent core, documenting each decision it makes. We can observe both the strengths and limitations reflected by this example.

\textbf{Initial Input.} The agent core receives the patient information along with a virtual image. For our evaluation task "Anomaly-wise Biomarker Calculation", we generate a corresponding QA pair based on the patient's complete medical record. The initial input is provided below:

\vspace{3pt}
\begin{lstlisting}[
    backgroundcolor=\color{lightgray!20},
    basicstyle=\small\ttfamily,
    breaklines=true,
    frame=single,
    breakatwhitespace=true,
    columns=flexible,
    label=lst:initial_input,
    caption=Initial Input
]
$Information$: {
    "Age": "62",
    "Sex": "Female",
    "Height": "165",
    "Weight": "72",
    "History": "Hypertension, osteoarthritis of the knees",
    "Complaint": "Chronic neck pain and stiffness, radiating pain to right shoulder"
},

$Query$: From an anomaly perspective in a specific medical image, after identifying 
the type and area, could you quantify specific biomarker characteristics?

$Image$: `PLACEHOLDER_IMAGE'
    
\end{lstlisting}

\textbf{Task Decomposition.} The agent core processes these inputs by analyzing the patient information and identifying key data for extraction. It determines the appropriate high-level tool chain required for task completion. The agent core then stores all relevant information in a memory bank and establishes a tool category chain to guide subsequent execution. The process is illustrated below:

\begin{lstlisting}[
    backgroundcolor=\color{lightgray!20},
    basicstyle=\small\ttfamily,
    breaklines=true,
    frame=single,
    breakatwhitespace=true,
    columns=flexible,
    label=lst:stage1_task_decomposition,
    caption=Stage 1: Task Decomposition
]

Initial Output: Known Info: []
Tool Chain: [*Anatomy Classifier* -> *Modality Classifier* -> 
*Anomaly Detector* -> *Biomarker Quantifier*]
Initial Value Dict: {'$Image$': 'PLACEHOLDER_IMAGE', '$Information$': 'PLACEHOLDER_
INFORMATION'}
Initial Score Dict: {'$Image$': 1.0, '$Information$': 1.0}
Initial Fixed Dict: frozendict.frozendict({'$Image$': 1.0, '$Information$': 1.0})
High-level Tool chain: Anatomy Classifier -> Modality Classifier -> 
Anomaly Detector -> Biomarker Quantifier

Memory bank: {'$Image$': 'PLACEHOLDER_IMAGE', '$Information$': 'PLACEHOLDER_
INFORMATION'} 
Score bank: {'$Image$': 1.0, '$Information$': 1.0} 
Fixed bank: frozendict.frozendict({'$Image$': 1.0, '$Information$': 1.0})
\end{lstlisting}

\noindent{Tool Selection \& Execution:} The first phase of tool selection and execution commences. At each step, the agent core integrates the planned high-level tool chain with stored information from the memory bank to generate a tool API call, specifying the purpose, tool name, and input parameters. The Platform then executes the API call by activating the selected tool and processing the input. Upon successful execution, the tool's output is stored in the memory bank. We configure the tool set according to the NR.~Denyl2 condition, where there is a mismatch between the anomaly detection tool and the provided $Image$ characteristics (Head and Neck X-ray) in terms of anatomy and modality. The established tool set is as follows:

\begin{lstlisting}[
    backgroundcolor=\color{lightgray!20},
    basicstyle=\small\ttfamily,
    breaklines=true,
    frame=single,
    breakatwhitespace=true,
    columns=flexible,
    label=Tool Set Description,
    caption=Tool Set Description
]

"TOOL1": {
    "Name": "TOOL1",
    "Category": "Anatomy Classifier",
    "Ability": "Determine the anatomy of the Image.",
    "Property": "Universal Anatomy Classifier",
    "Compulsory Input": [
        "$Image$"
    ],
    "Optional Input": [],
    "Output": [
        "$Anatomy$"
    ],
    "Performance": "Score from 0.95 to 0.95, increases with optional inputs"
},
"TOOL2": {
    "Name": "TOOL2",
    "Category": "Modality Classifier",
    "Ability": "Determine the modality of the Image.",
    "Property": "Universal Modality Classifier",
    "Compulsory Input": [
        "$Image$"
    ],
    "Optional Input": [],
    "Output": [
        "$Modality$"
    ],
    "Performance": "Score from 0.95 to 0.95, increases with optional inputs"
},

...

"TOOL5": {
    "Name": "TOOL5",
    "Category": "Anomaly Detector",
    "Ability": "Given the Limb Ultrasound Image, determine the location and 
    type of abnormality.",
    "Property": "Anomaly Detector only suitable for Limb Ultrasound image",
    "Compulsory Input": [
        "$Image$"
    ],
    "Optional Input": [],
    "Output": [
        "$AnomalyMask$",
        "$AnomalyObject$"
    ],
    "Performance": "Score from 0.8 to 0.8, increases with optional inputs"
},
"TOOL6": {
    "Name": "TOOL6",
    "Category": "Anomaly Detector",
    "Ability": "Given the Breast MRI Image, determine the location and type 
    of abnormality.",
    "Property": "Anomaly Detector only suitable for Breast MRI image",
    "Compulsory Input": [
        "$Image$"
    ],
    "Optional Input": [],
    "Output": [
        "$AnomalyMask$",
        "$AnomalyObject$"
    ],
    "Performance": "Score from 0.8 to 0.8, increases with optional inputs"
},

...

"TOOL8": {
    "Name": "TOOL8",
    "Category": "Imaging Diagnoser",
    "Ability": "Given the Head and Neck X-ray Image, diagnose the disease.",
    "Property": "Imaging Diagnoser only suitable for Head and Neck X-ray image",
    "Compulsory Input": [
        "$Image$"
    ],
    "Optional Input": [],
    "Output": [
        "$Disease$"
    ],
    "Performance": "Score from 0.75 to 0.75, increases with optional inputs"
},

...

"TOOL10": {
    "Name": "TOOL10",
    "Category": "Biomarker Quantifier",
    "Ability": "Measure the anomaly biomarker of the Image.",
    "Property": "Universal Anomaly Biomarker Quantifier",
    "Compulsory Input": [
        "$Image$",
        "$AnomalyObject$",
        "$AnomalyMask$"
    ],
    "Optional Input": [
        "$AnomalyDim$"
    ],
    "Output": [
        "$AnomalyDim$",
        "$AnomalyQuant$"
    ],
    "Performance": "Score from 0.75 to 0.8, increases with optional inputs"
},

...
        
\end{lstlisting}

In this initial step, the agent core selects the anatomy classification tool. We observe that the anatomical information is successfully stored in the memory bank after the tool execution:

\begin{lstlisting}[
    backgroundcolor=\color{lightgray!20},
    basicstyle=\small\ttfamily,
    breaklines=true,
    frame=single,
    breakatwhitespace=true,
    columns=flexible,
    label=step1,
    caption=Tool Selection \& Execution Step 1
]

Step 1 starts!
Stepoutput: Based on the tool chain from Part 1 (Anatomy Classifier -> 
Modality Classifier -> Anomaly Detector -> Biomarker Quantifier) and current 
results dictionary, I'll proceed with the first step:

<Call>
    <Purpose>Determine the anatomy region of the medical image</Purpose>
    <Tool>TOOL1</Tool>
    <Input>['$Image$']</Input>
</Call>
Call Dict: {'Purpose': 'Determine the anatomy region of the medical image', 'Tool': 
'TOOL1', 'Input': ['$Image$']}
Memory bank: {'$Image$': 'PLACEHOLDER_IMAGE', '$Information$': 'PLACEHOLDER_
INFORMATION', '$Anatomy$': 'Head and Neck'} 
Score bank: {'$Image$': 1.0, '$Information$': 1.0, '$Anatomy$': 0.95} 
Fixed bank: frozendict.frozendict({'$Image$': 1.0, '$Information$': 1.0})
Step 1 completes!

\end{lstlisting}

In the second step, the modality classification tool is selected and executed. As shown below, the resulting modality information is added to the memory bank.

\begin{lstlisting}[
    backgroundcolor=\color{lightgray!20},
    basicstyle=\small\ttfamily,
    breaklines=true,
    frame=single,
    breakatwhitespace=true,
    columns=flexible,
    label=step2,
    caption=Tool Selection \& Execution Step 2
]

Step 2 starts!
Stepoutput: Based on the tool chain and current results showing Head and Neck anatomy, 
the next step is to determine the image modality:

<Call>
    <Purpose>Determine the imaging modality of the medical image</Purpose>
    <Tool>TOOL2</Tool>
    <Input>['$Image$']</Input>
</Call>
Call Dict: {'Purpose': 'Determine the imaging modality of the medical image', 'Tool': 
'TOOL2', 'Input': ['$Image$']}
Memory bank: {'$Image$': 'PLACEHOLDER_IMAGE', '$Information$': 'PLACEHOLDER_
INFORMATION', '$Anatomy$': 'Head and Neck', '$Modality$': 'X-ray'} 
Score bank: {'$Image$': 1.0, '$Information$': 1.0, '$Anatomy$': 0.95, '$Modality$': 
0.95} 
Fixed bank: frozendict.frozendict({'$Image$': 1.0, '$Information$': 1.0})
Step 2 completes!

\end{lstlisting}

In the third step, contrary to the expected utilization of the organ segmentation tool, the disease diagnosis tool is invoked instead. While this tool selection is irrelevant to the current task, the execution completes without errors. The tool call is detailed below:

\begin{lstlisting}[
    backgroundcolor=\color{lightgray!20},
    basicstyle=\small\ttfamily,
    breaklines=true,
    frame=single,
    breakatwhitespace=true,
    columns=flexible,
    label=step3,
    caption=Tool Selection \& Execution Step 3]
Step 3 starts!
Stepoutput: Given the current results showing Head and Neck anatomy and X-ray modality, 
the next step is to detect any anomalies in the medical image:

<Call>
    <Purpose>Detect abnormalities in the Head and Neck X-ray image</Purpose>
    <Tool>TOOL8</Tool>
    <Input>['$Image$']</Input>
</Call>
Call Dict: {'Purpose': 'Detect abnormalities in the Head and Neck X-ray image', 'Tool': 
'TOOL8', 'Input': ['$Image$']}
Memory bank: {'$Image$': 'PLACEHOLDER_IMAGE', '$Information$': 'PLACEHOLDER_
INFORMATION', '$Anatomy$': 'Head and Neck', '$Modality$': 'X-ray', '$Disease$': 
'Cervical spine degenerative changes'} 
Score bank: {'$Image$': 1.0, '$Information$': 1.0, '$Anatomy$': 0.95, '$Modality$': 
0.95, '$Disease$': 0.75} 
Fixed bank: frozendict.frozendict({'$Image$': 1.0, '$Information$': 1.0})
Step 3 completes!

\end{lstlisting}

During the fourth step, when selecting the anomaly detection tool, the agent core correctly identifies that the available tools in the set lack the capability to detect and localize anomalies in Head and Neck X-ray images. As demonstrated below, the agent core appropriately rejects this invalid operation and successfully recognizes the tools' capability gap.

\vspace{3pt}
\begin{lstlisting}[
    backgroundcolor=\color{lightgray!20},
    basicstyle=\small\ttfamily,
    breaklines=true,
    frame=single,
    breakatwhitespace=true,
    columns=flexible,
    label=step4,
    caption=Tool Selection \& Execution Step 4]

Step 4 starts!
Stepoutput: Before we can quantify biomarkers (per the original tool chain), we need 
the anomaly detection results first. Looking at the available tools for Head and Neck 
X-ray anomaly detection:

<NoCall>
    <Purpose>Detect specific anomalies in Head and Neck X-ray for biomarker 
    quantification</Purpose>
    <Category>Anomaly Detector</Category>
    <Anatomy>Head and Neck</Anatomy>
    <Modality>X-ray</Modality>
    <Ability>SpecificToolMissing</Ability>
</NoCall>
Deny Call Dict: {'Purpose': 'Detect specific anomalies in Head and Neck X-ray for 
biomarker quantification', 'Category': 'Anomaly Detector', 'Anatomy': 'Head and Neck', 
'Modality': 'X-ray', 'Ability': 'SpecificToolMissing'}Agent denies the call because no 
suitable tool is found.

\end{lstlisting}

Overall, while the agent core demonstrates correct initialization and tool chain planning, it exhibits a deviation during the tool selection \& execution stage. Despite having access to tool chain guidance in each round, the agent core fails to follow these directives, resulting in an unnecessary disease diagnosis tool call. Nevertheless, the execution proceeds without any IO errors, and the agent core successfully identifies the tools' capability limitations, appropriately concluding with a denial of task execution due to the missing functionality.

\end{document}